\documentclass{article}

\usepackage{microtype}
\usepackage{graphicx}
\usepackage{booktabs} 

\usepackage{hyperref}

\usepackage[accepted]{icml2024}

\usepackage{amsmath}
\usepackage{amssymb}
\usepackage{mathtools}
\usepackage{amsthm}

\usepackage[capitalize,noabbrev]{cleveref}

\theoremstyle{plain}
\newtheorem{theorem}{Theorem}[section]

\theoremstyle{definition}
\newtheorem{definition}[theorem]{Definition}

\theoremstyle{remark}

\usepackage{caption}
\usepackage{subcaption}
\usepackage{float}
\usepackage{xcolor}
\definecolor{mygray}{gray}{0.7}

\usepackage[Base]{optional} 

\usepackage{listings}
\usepackage{glossaries}
\glsdisablehyper
\newacronym{cae}{CAE}{Concrete Autoencoder}
\newacronym{ip}{IP}{Indirect Parametrization}
\newacronym{gjsd}{GJSD}{Generalized Jensen-Shannon Divergence}
\newacronym{stg}{STG}{Stochastic Gaussian Gate}
\newacronym{wast}{WAST}{Where to pay Attention during Sparse Training}

\DeclareMathOperator*{\argmax}{arg\,max}

\usepackage{mathtools}
\DeclarePairedDelimiterX{\infdivx}[2]{(}{)}{%
  #1\;\delimsize\|\;#2%
}

\DeclarePairedDelimiter{\norm}{\lVert}{\rVert}

\usepackage{multirow}
\usepackage{tabularx}

\icmltitlerunning{Indirectly Parameterized Concrete Autoencoders}

\begin{document}

\twocolumn[
\icmltitle{Indirectly Parameterized Concrete Autoencoders}


\icmlsetsymbol{equal}{*}

\begin{icmlauthorlist}
\icmlauthor{Alfred Nilsson}{equal,kth,scilife}
\icmlauthor{Klas Wijk}{equal,kth,serc}
\icmlauthor{Sai bharath chandra Gutha}{kth}
\icmlauthor{Erik Englesson}{kth}
\icmlauthor{Alexandra Hotti}{kth,scilife,klarna}
\icmlauthor{Carlo Saccardi}{kth}
\icmlauthor{Oskar Kviman}{kth,scilife}
\icmlauthor{Jens Lagergren}{kth,scilife}
\icmlauthor{Ricardo Vinuesa}{kth,serc}
\icmlauthor{Hossein Azizpour}{kth,serc}
\end{icmlauthorlist}

\icmlaffiliation{kth}{KTH Royal Institute of Technology, Stockholm, Sweden}
\icmlaffiliation{serc}{Swedish e-Science Research Centre (SeRC), Stockholm, Sweden}
\icmlaffiliation{scilife}{Science for Life Laboratory, Solna, Sweden}
\icmlaffiliation{klarna}{Klarna, Stockholm, Sweden}

\icmlcorrespondingauthor{Alfred Nilsson}{alfredn@kth.se}
\icmlcorrespondingauthor{Klas Wijk}{kwijk@kth.se}

\icmlkeywords{Feature Selection, Differentiable Optimization, Gumbel-Softmax, Concrete Autoencoders}

\vskip 0.3in
]



\printAffiliationsAndNotice{\icmlEqualContribution} 

\begin{abstract}
Feature selection is a crucial task in settings where data is high-dimensional or acquiring the full set of features is costly. Recent developments in neural network-based embedded feature selection show promising results across a wide range of applications. Concrete Autoencoders (CAEs), considered state-of-the-art in embedded feature selection, may struggle to achieve stable joint optimization, hurting their training time and generalization. In this work, we identify that this instability is correlated with the CAE learning duplicate selections. To remedy this, we propose a simple and effective improvement: Indirectly Parameterized CAEs (IP-CAEs). IP-CAEs learn an embedding and a mapping from it to the Gumbel-Softmax distributions' parameters. Despite being simple to implement, IP-CAE exhibits significant and consistent improvements over CAE in both generalization and training time across several datasets for reconstruction and classification. Unlike CAE, IP-CAE effectively leverages non-linear relationships and does not require retraining the jointly optimized decoder. Furthermore, our approach is, in principle, generalizable to Gumbel-Softmax distributions beyond feature selection.
\end{abstract}
\section{Introduction} \label{sec:introduction}
Feature selection is a fundamental task in machine learning and statistics, enabling more parsimonious and interpretable models. It is essential in several applications such as bioinformatics e.g., gene subset selection, neuroscience e.g., fMRI analysis, and fluid mechanics e.g., optimal sensor placement. Moreover, feature selection is often used for regularization. Unfortunately, finding the optimal selection is NP-hard \citep{amaldi_approximability_1998}.

Although a large body of work exists on feature selection~\citep{cai2018feature}, due to the success of deep networks, neural network-based embedded feature selection has gained more interest~\citep{balin_concrete_2019, yamada_feature_2020, lemhadri_lassonet_2021}. Among those, \Glspl{cae} \citep{balin_concrete_2019}, is an established approach which allows for differentiable feature selection using a layer consisting of stochastic Gumbel-Softmax distributed nodes \citep{maddison_concrete_2017, jang_gumbel_softmax_2017}. 

\begin{figure}[t]
    \centering
    \includegraphics[width=\linewidth]{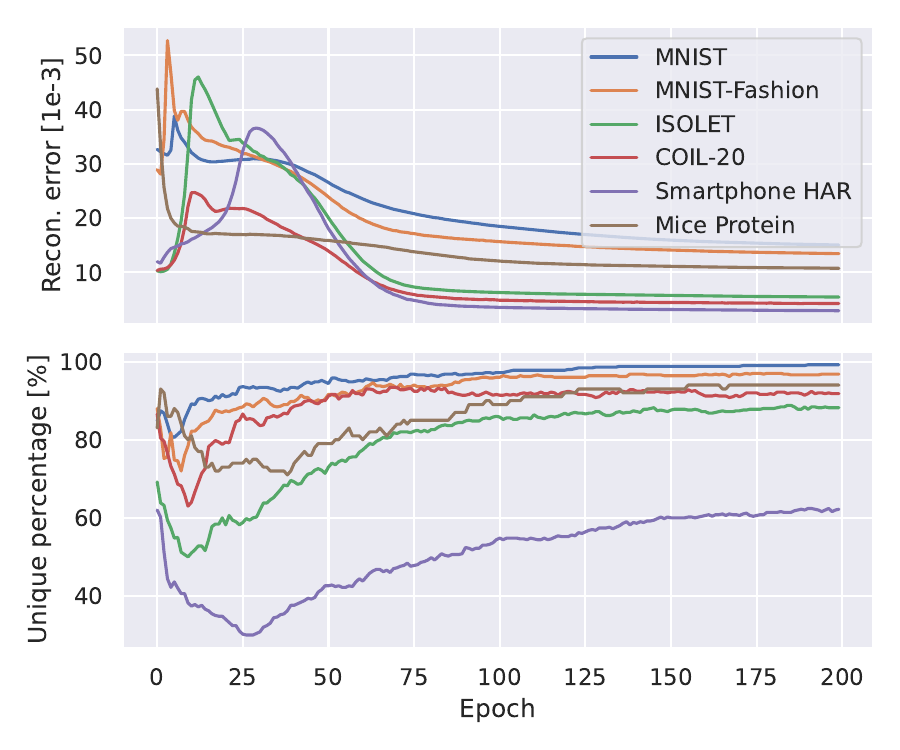}
    \caption{\textbf{\gls{cae} Training Instability}. For most datasets, the \gls{cae} architecture exhibits a large spike in reconstruction error that consistently correlates with the unique percentage (definition \ref{def:unique-percentage}).}
    \label{fig:teaser}
\end{figure}

\begin{figure}[t]
    \centering
    \begin{subfigure}[b]{\linewidth}
         \centering
         \includegraphics[width=0.72\linewidth]{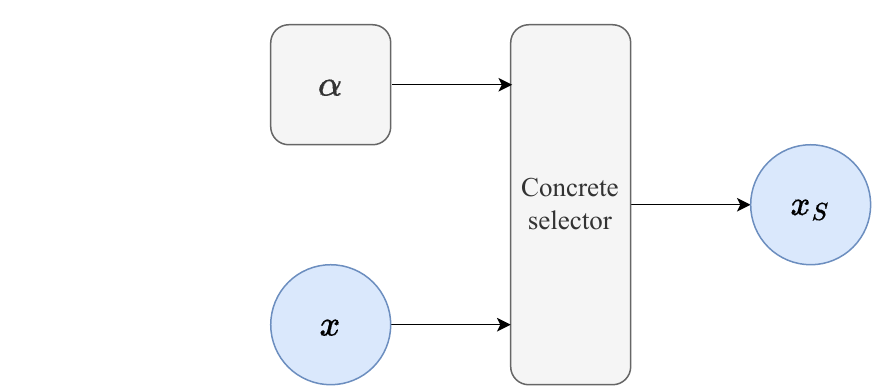}
         \caption{Direct parametrization}
         \label{fig:direct}
     \end{subfigure}
     \begin{subfigure}[b]{\linewidth}
         \vspace{10pt}
         \centering
         \includegraphics[width=0.72\linewidth]{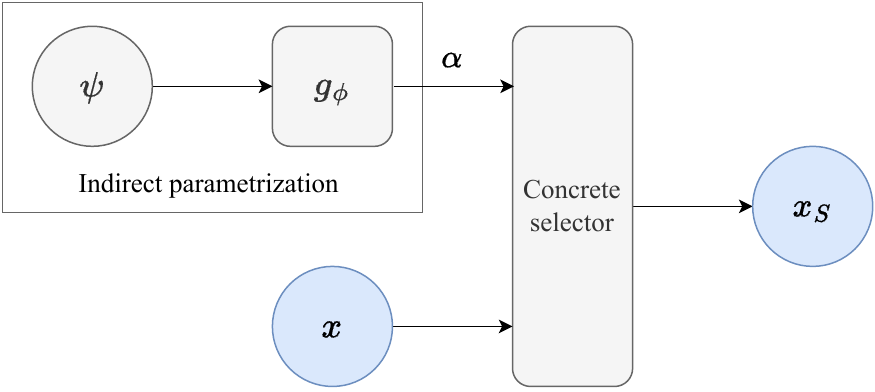}
         \caption{Indirect parametrization}
         \label{fig:indirect}
     \end{subfigure}
     \caption{\textbf{Architecture}. An overview of the \gls{cae} architecture, showcasing \gls{ip}. Instead of directly learning $\alpha$, we propose to learn an embedding $\psi$ and a transformation $g_\phi$ that output $\alpha$.}
     \label{fig:architecture}
\end{figure}

In this work, we identify a recurring instability issue of \Glspl{cae} (\cref{fig:teaser}, top) which leads to increased training time and subpar performance. We then show that the instability correlates with selection of redundant features (\cref{fig:teaser}, bottom). 
To remedy this, we propose a simple modification to indirectly parametrize the Gumbel-Softmax distributions via a learnable embedding and transformation (\cref{fig:architecture}), we refer to this alternative as Indirectly Parameterized CAEs (IP-CAEs) and rigorously verify their empirical effectiveness. We summarize this paper's main contributions below.
\begin{itemize}
    \item We identify training instability in CAE and show it strongly correlates with redundant features (\cref{fig:teaser}).
    \item We introduce IP-CAE (\cref{fig:architecture}), a simple and effective way to alleviate the instability of vanilla CAE (\cref{fig:recon-curves} solid lines), and show it leads to unique selections (\cref{fig:recon-curves} dotted lines), improved accuracy (\cref{fig:acc-curves}) and training time (\cref{tab:speedup}). We also study the update rules of CAE and IP-CAE, showing the latter learns a transformation of gradients (\cref{sec:method:update_rule}).
    \item We propose and compare against \gls{gjsd} regularization 
    (\cref{sec:method:gjsd})
    , an explicit, probabilistic approach to mitigate duplicate selections, and show while \gls{gjsd} regularization is effective, IP-CAE is superior (Tables~\ref{tab:reconstruction} and \ref{tab:classification}).
    \item We demonstrate successful end-to-end training of \gls{cae} architectures for both reconstruction (\cref{tab:reconstruction}) and classification (\cref{tab:classification}) achieving state-of-the-art results on multiple datasets.
    \item We study the various aspect of IP-CAE and specifically show that it does not require additional hyperparameter tuning and that its superior performance is insensitive to the number of selected features (\cref{fig:varying-k}) and the size of indirect parametrization (\cref{fig:varying-P}).
\end{itemize}
\section{Method} \label{sec:method}

In this section we describe the vanilla CAE, discuss its shortcomings and present our proposed improvements to \gls{cae} training.
We introduce a new way to \textit{indirectly} parameterize the Gumbel-Softmax distributions in CAE, which we call \glsfirst{ip}.
We further propose a baseline diversity-encouraging regularization method using the Generalized Jensen Shannon Divergence (\gls{gjsd}).

\subsection{Concrete Autoencoders (CAE)}
The \gls{cae} \citep{balin_concrete_2019} architecture is competitive with state-of-the-art methods in neural network-based embedded feature selection. \glspl{cae} consist of two components: a concrete selection layer that performs differentiable feature selection on the input features (encoder) and an arbitrary neural network (decoder). A predictive network is used in place of the decoder for classification or regression tasks.

The concrete selector layer consists of $K$ independent Gumbel-Softmax distributed variables $\boldsymbol{m}_j$ \citep{jang_gumbel_softmax_2017, maddison_concrete_2017}:  
\begin{equation}
    \boldsymbol{m}_j = \frac{\exp\{ (\log\boldsymbol{\alpha}_j + \boldsymbol{g}_j) / T\}}{\sum_{i=1}^D \exp\{ (\log \boldsymbol{\alpha}_{j,i} + \boldsymbol{g}_{j,i}) / T\}},
\end{equation}
where $\log\boldsymbol{\alpha}_j \in \mathbb{R}^D$ for $j \in 1, 2, \hdots, K$ are the distributions parameters (logits), $\boldsymbol{g}_j \in \mathbb{R}^D$ are i.i.d. standard Gumbel distributed \citep{gumbelMaximaMeanLargest1954}, and $T \in \mathbb{R}_+$ is a global temperature that is annealed throughout the training.

The samples are multiplied with the input features and passed through the decoder network. Using the reparametrization trick, the parameters are learnable through backpropagation from the decoder network's output. 
By forming a matrix whose rows contain $\{ \mathbf{m}_j \}_{j=1}^{k}$ and denoting it by $\boldsymbol{M} \in \mathbb{R}^{K \times D}$, we can express the complete subset selection according to CAE as:
\begin{equation}
    \mathbf{x}_S = \boldsymbol{M}\mathbf{x},
\end{equation}
where $\mathbf{x}_S \in \mathbb{R}^{K}$. Then, the selected features serve as the input to an arbitrary neural network $f_{\theta}$ the output of which is used to calculate a loss. MSE and cross-entropy losses are commonly used for reconstruction and classification respectively.

\citet{balin_concrete_2019} propose exponential annealing from a starting temperature $T_0$ to a final temperature $T_B$ according to the following annealing schedule which we also use:
\begin{equation}
    T(b) = T_0 \left ( \frac{T_B}{T_0} \right )^\frac{b}{B},
\end{equation}
where $b \in \mathbb{N}$ is the current epoch and $B \in \mathbb{N}$ is the total number of epochs. The authors find that this schedule works for a broad range of datasets and is not sensitive to the specific start and end temperatures chosen.

As $T \rightarrow 0$, the Gumbel-Softmax samples $\boldsymbol{m}_j$ approach one-hot vectors corresponding to single input features. At test time, we evaluate the decoder using \emph{discrete} input features selected according to $\argmax_j \log\boldsymbol{\alpha}_{i, j}$, where $i \in \{ 1, 2, \hdots, K \} = [K]$.

\subsection{Challenges of CAE}
\label{sec:CAE_challenges}
In principle, it is not guaranteed that the learned parameters of the Gumbel-Softmax will correspond to distinct input features at any given point during training. We quantify the diversity of the Gumbel-Softmax parameters using the Unique Percentage (UP).


\begin{definition}[Unique Percentage]
\label{def:unique-percentage}
For a given set of Gumbel-Softmax parameters (logits) $\log\boldsymbol{\alpha} \in \mathbb{R}^{K \times D}$:
\begin{equation}
    \mathit{UP}(\boldsymbol{\alpha}) = 100 \cdot \frac{|\{ \argmax_j \log\boldsymbol{\alpha}_{i, j} : i \in [K] \}|}{K},
\end{equation}
is the percentage of unique maximum parameter indices. Note that $D$ denotes the total number of features and $K$ the number of selected features.
\end{definition}

We empirically demonstrate instability during training of \glspl{cae} (\cref{fig:teaser}, top). Interestingly, our results show that this instability strongly correlates with the unique percentage, consistently across tasks and datasets (\cref{fig:teaser}, bottom). 

Furthermore, we empirically establish three more shortcomings of CAEs: (i) a large number of training epochs is required for CAE to converge to a local minimum, (ii) the quality of these local minima sometimes exhibit a high variance, and (iii) an end-to-end optimization of a non-linear decoder might incur additional instability, especially for prediction tasks other than reconstruction.

In the next subsection we describe our proposed IP-CAE which is later shown to alleviate the aforementioned shortcoming of vanilla CAE including instability, unique percentange, variance (\cref{fig:training-curves}), training time (\cref{tab:speedup}), and non-linear decoder (\cref{fig:hidden}) leading to state-of-the-art performance for feature selection for both reconstruction (\cref{tab:reconstruction}) and classification (\cref{tab:classification}) on all datasets considered.


\subsection{Indirect Parametrization}

We investigate parameterizing $\log\boldsymbol{\alpha} \in \mathbb{R}^{K \times D}$ by transforming an array of learnable parameters $\boldsymbol{\Psi} \in \mathbb{R}^{K \times P}$ with a network 
$g_\phi$ as follows:
\begin{equation}
    \log\boldsymbol{\alpha}_i = g_\phi(\boldsymbol{\psi}_i),
\end{equation}
where $\log \boldsymbol{\alpha}_i \in \mathbb{R}^{D}$ and $\boldsymbol{\psi}_i \in \mathbb{R}^{P}$ are the transposed $i$th rows of $\log \boldsymbol{\alpha}$ and $\Psi$, respectively.
In our experiments, 
we let $g_\phi$ be a linear network that is shared across stochastic nodes, \textit{i.e.} 
 $\phi = (\boldsymbol{W}, \boldsymbol{b})$ 
and: 
\begin{equation}
\label{eq:ip}
    \log \boldsymbol{\alpha}_i = \boldsymbol{W}\boldsymbol{\psi}_i + \boldsymbol{b}
    , \quad i \in [K],
\end{equation}
where $\boldsymbol{W} \in \mathbb{R}^{D \times P}$ and $\boldsymbol{b} \in \mathbb{R}^D$.
This can be interpreted as a feature embedding with embedding dimensionality $P$.

\subsection{A Comparison Between CAE and IP-CAE}
\label{sec:method:update_rule}
As the difference between CAE and IP-CAE is in how $\log \boldsymbol{\alpha}$ is parameterized, we analyze the differences between the methods by studying the following update rule based on different parameterizations: $\log \boldsymbol{\alpha}_i^{(t+1)}~\leftarrow~\log \boldsymbol{\alpha}_i^{(t)}~-~\eta \nabla \mathcal{L}$, where $\mathcal{L}$ is the gradient of the loss with respect to $\log \boldsymbol{\alpha}_i^{(t)}$.

\Gls{cae} can be interpreted as a trivial case of IP, where $\log{\boldsymbol{\alpha}}$ is directly parameterized by a learnable buffer $\boldsymbol{\Psi} \in \mathbb{R}^{K \times P}$ with $P=D$, such that $\log{\boldsymbol{\alpha}}_i = \boldsymbol{\psi}_i$. In this case, the update rule for $\log{\boldsymbol{\alpha}}_i$ is:
\begin{align}
\log \boldsymbol{\alpha}_i^{(t+1)} \leftarrow \boldsymbol{\psi}_i - \eta \nabla \mathcal{L}
\end{align}
where $t$ is the current optimization step, $\eta$ is the learning rate, and $\nabla \mathcal{L} \in \mathbb{R}^{D}$ is the gradient of the loss function $\mathcal{L}$ with respect to $\log \boldsymbol{\alpha}_i^{(t)} = \boldsymbol{\psi}_i^{(t)} = \boldsymbol{\psi}_i$. 

For simplicity, we consider IP-CAE with $P=D$ and without the bias term, and thus have  $\log{\boldsymbol{\alpha}}_i = \boldsymbol{W}\boldsymbol{\psi}_i$, with the following update rule (detailed derivations in \cref{sec:appendix:update_rules}):
\begin{align}
&\log \boldsymbol{\alpha}_i^{(t+1)} \leftarrow \boldsymbol{W}\boldsymbol{\psi}_i - \eta \boldsymbol{T}_i \nabla \mathcal{L} \\ 
&
\label{eq:ip_update_rule}
\boldsymbol{T}_i = \boldsymbol{W} \boldsymbol{W}^T + \boldsymbol{\psi}_i^T(\boldsymbol{\psi}_i - \eta \boldsymbol{W}^T \nabla \mathcal{L}) \boldsymbol{I}
\end{align}
where $\nabla \mathcal{L}$ is the gradient of the loss with respect to $\log \boldsymbol{\alpha}_i^{(t)} = \boldsymbol{W}^{(t)}\boldsymbol{\psi}_i^{(t)} = \boldsymbol{W}\boldsymbol{\psi}_i$, and the step-dependent matrix $\boldsymbol{T}_i \in \mathbb{R}^{D \times D}$ represents a learned transformation of the gradients. The transform affects the gradients in two ways: a linear transformation represented by $\boldsymbol{W}\boldsymbol{W}^T$ that is shared for all $i$, and a scaling by $\boldsymbol{\psi}_i^T(\boldsymbol{\psi}_i - \eta \boldsymbol{W}^T \nabla \mathcal{L}) \in \mathbb{R}$, which is the dot product between $\boldsymbol{\psi}_i^{(t)}$ and $\boldsymbol{\psi}_i^{(t+1)}$. A geometric interpretation of the dot product is $||\boldsymbol{\psi}_i^{(t)}||_2 ||\boldsymbol{\psi}_i^{(t+1)}||_2 \cos(\theta_i)$ with $\theta_i$ being the angle between the two vectors. 
Empirically, we have found that the learned rescaling changes throughout training, and our results suggest that these changes are beneficial. Because of the interactions between $\boldsymbol{W}$, $\phi$, and $\nabla\mathcal{L}$, the effect throughout training may be elaborate. We include additional experiments exploring the update's behavior in \cref{app:gradients}.

Thus, simply changing the parameterization of $\log \boldsymbol{\alpha}_i$ from $\boldsymbol{\psi}_i$ to $\boldsymbol{W}\boldsymbol{\psi}_i$ significantly changes the update rule to transform all gradients by $\boldsymbol{W}\boldsymbol{W}^T$ and scale specific gradients by the dot product between the current and next step of $\boldsymbol{\psi}_i$. Clearly, IP-CAE is a generalization of CAE, as CAE corresponds to the special case with a fixed $\boldsymbol{W}=\boldsymbol{I}$.

\subsection{Generalized Jensen-Shannon Divergence}
\label{sec:method:gjsd}
Since the aforementioned challenges of CAE (\cref{sec:CAE_challenges}), particularly its instability, is strongly correlated with reduced unique percentage (\cref{fig:teaser}), we propose a direct mechanism to encourage diversity of selected features by using the GJSD as a regularization, which will serve as an important baseline in our empirical study. We later show its effectiveness in comparison to the vanilla CAE but IP remains superior (\cref{tab:classification,,tab:reconstruction}).

\begin{definition}[Generalized Jensen--Shannon Div.]
\label{def:gjsd}
The \gls{gjsd} for $K$ categorical distributions $\{\boldsymbol{p}_i\}_{i=1}^{K}$, and weights $\mathbf{w}$ is given by:
\begin{equation}
D_{GJS}(\{\boldsymbol{p}_i\}_{i=1}^{K}) = \sum_{i=1}^{K} w_i D_{KL} \bigg ( \boldsymbol{p}_i \;||\;  \sum_{j=1}^{K} w_j  \boldsymbol{p}_j \bigg ),
\end{equation}
where $D_{KL}$ denotes the Kullback--Leibler divergence.
\end{definition}

\gls{gjsd} has previously been employed to measure the diversity among the mixture components \citep{pmlr-v151-kviman22a,pmlr-v202-kviman23a} and that can be utilized as a loss function \citep{englesson2021generalized, hendrycks2019augmix}. As the goal is to learn distinct Gumbel-Softmax distributions that converge to unique features, maximizing the $D_{GJS}$ can help prevent duplicate selections by encouraging diverse distributions. While we are concerned with Gumbel-Softmax distributions, they can be approximately treated as categoricals. We exploit this by calculating an approximate \gls{gjsd} with the probability vectors $S(\log \boldsymbol{\alpha}_i)$ of our Gumbel-Softmax distributions, where $S(\boldsymbol{z})=\exp{\{[z_1, \dots, z_K]}\}/ \sum_{i=1}^K \exp{\{z_i\}}$ is the softmax function. We assume equal weights for the mixture components, \textit{i.e.} $w_i \equiv 1 / K$.

Using definition \ref{def:gjsd}, we define our regularized loss function 
\begin{equation}
     \mathcal{L}_\lambda(\cdot, \log \boldsymbol{\alpha}) = \mathcal{L}(\cdot)  - \lambda D_{GJS}(\{S(\log  \boldsymbol{\alpha}_i)\}_{i=1}^{K}),
\end{equation}
where $\lambda > 0$ is a parameter controlling the regularization and $\mathcal{L}(\cdot)$ is the non-regularized loss (e.g. MSE for reconstruction and cross-entropy for classification).

\section{Related Work} \label{sec:related-work}
Feature selection methods are broadly categorized into three paradigms: filter methods, wrapper methods, and embedded methods \citep{feature_selection_survey}. While filter methods treat each feature independently and do not account for interactions between them, wrapper methods select features based on a black-box model. Finally, embedded methods perform feature selection as part of the model, usually through learning the feature selection throughout fitting the model. 

Next we describe recent state-of-the-art methods used for neural network-based embedded feature selection. 

\paragraph{Feature selection using STGs} \citet{yamada_feature_2020}, similarly to \gls{cae}, use stochastic nodes to sample features during training to perform differentiable joint optimization of feature selection and model. While \gls{cae} models the selection using $K$ categorical nodes in the concrete selector layer, \glspl{stg} uses $D$ Bernoulli nodes, each for one input feature, where $K$ denotes the desired number of optimal features, and $D$ denotes the total number of input features. The authors propose a novel reparametrization for Bernoulli variables using thresholded Gaussian variables to allow for differentiable learning. The use of Bernoulli gates is closely related to the Bernoulli-Gaussian model for linear regression with feature selection and $\ell_0$ regularization.

\paragraph{LassoNet} \citet{lemhadri_lassonet_2021} extend the popular Lasso \citep{tibshiraniRegressionShrinkageSelection1996} method for regression. Although the classic Lasso has been efficient and useful for embedded feature selection in linear models, it is challenging to generalize it to neural network models \citep{CUI2016505}. The LassoNet architecture achieves this by introducing residual connections from the input layer to the output of the network and applying an $\ell_1$ penalty term to that layer. The design principle is such that it allows for a feature to be selected by the model if and only if it also gets selected by the residual layer. The authors model this principle as an explicit constraint in the optimization problem and propose a projected proximal gradient optimization algorithm to ensure the constraint satisfiability during the process.

\begin{table*}[ht]
    \caption{\textbf{Reconstruction Error}. Mean normalized Frobenius norm for the reconstruction task. The values are an average of 10 repetitions $\pm$ 1 standard deviation.}
    \label{tab:reconstruction}
    \begin{center}
    \begin{small}
        \begin{sc}
            \begin{tabular}{{p{2.1cm}}*{6}{p{1.8cm}}}
            Model                     & MNIST                                     & MNIST-Fashion                             & ISOLET                                    & COIL-20                                   & Smartphone HAR                            & Mice \allowbreak Protein \\
            \hline
            STG                       & 2.42E-02  \textcolor{mygray}{$\pm$ 2.70E-06} & 1.80E-02  \textcolor{mygray}{$\pm$ 2.84E-06}& 7.00E-03  \textcolor{mygray}{$\pm$ 1.96E-06}& 5.00E-03  \textcolor{mygray}{$\pm$ 6.75E-06}& 4.00E-03  \textcolor{mygray}{$\pm$ 3.04E-06}& 1.17E-02  \textcolor{mygray}{$\pm$ 6.43E-05}\\
            LassoNet                  & 1.98E-02  \textcolor{mygray}{$\pm$ 9.02E-06} & 1.96E-02  \textcolor{mygray}{$\pm$ 4.23E-06}& 7.40E-03  \textcolor{mygray}{$\pm$ 4.25E-06}& 6.24E-03  \textcolor{mygray}{$\pm$ 5.42E-06}& 4.01E-03  \textcolor{mygray}{$\pm$ 1.29E-06}& 1.16E-02  \textcolor{mygray}{$\pm$ 4.45E-05}\\
            CAE                       & 1.48E-02  \textcolor{mygray}{$\pm$ 1.31E-04} & 1.36E-02  \textcolor{mygray}{$\pm$ 1.17E-04}& 5.42E-03  \textcolor{mygray}{$\pm$ 6.25E-05}& 4.21E-03  \textcolor{mygray}{$\pm$ 7.69E-05}& 3.03E-03  \textcolor{mygray}{$\pm$ 6.45E-05}& 8.71E-03  \textcolor{mygray}{$\pm$ 2.61E-04}\\
            \gls{gjsd}                & 1.45E-02  \textcolor{mygray}{$\pm$ 1.14E-04} & 1.23E-02  \textcolor{mygray}{$\pm$ 9.26E-05}& 4.86E-03  \textcolor{mygray}{$\pm$ 4.52E-05}& 2.80E-03  \textcolor{mygray}{$\pm$ 2.85E-05}& 2.57E-03  \textcolor{mygray}{$\pm$ 3.50E-05}& 8.23E-03  \textcolor{mygray}{$\pm$ 2.99E-04}\\
            IP-CAE                    & \textbf{1.36E-02}  \textcolor{mygray}{$\pm$ \textbf{5.35E-05}} & \textbf{1.17E-02}  \textcolor{mygray}{$\pm$ \textbf{3.15E-05}}& \textbf{4.38E-03}  \textcolor{mygray}{$\pm$ \textbf{1.28E-05}}& \textbf{2.48E-03}  \textcolor{mygray}{$\pm$ \textbf{1.39E-05}}&\textbf{2.03E-03}  \textcolor{mygray}{$\pm$ \textbf{2.75E-05}}& \textbf{6.90E-03}  \textcolor{mygray}{$\pm$ \textbf{1.25E-04}}\\
            \end{tabular}
        \end{sc}
    \end{small}
\end{center}
\end{table*}

\begin{table*}[ht]
    \caption{\textbf{Classification Accuracy}. Top-1 accuracy for the classification task. The values are an average of 10 repetitions $\pm$ 1 standard deviation.}
    \label{tab:classification}
    \begin{center}
    \begin{small}

            \begin{tabular}{{p{2.4cm}}*{6}{p{1.8cm}}}
            Model                   & MNIST                     & MNIST-Fashion             & ISOLET                    & COIL-20                   & Smartphone HAR            & Mice \allowbreak Protein \\
            \hline
            STG                     & $92.29 \;\;\textcolor{mygray}{\pm 0.30}$          & $80.85 \;\;\textcolor{mygray}{\pm 0.27}$          & $84.95 \;\;\textcolor{mygray}{\pm 0.31}$          & $96.80 \;\;\textcolor{mygray}{\pm 0.25}$          & $88.80 \;\;\textcolor{mygray}{\pm 0.08}$          & $68.24 \;\;\textcolor{mygray}{\pm 1.11}$ \\
            LassoNet                & $90.06 \;\;\textcolor{mygray}{\pm 0.33}$          & $78.28 \;\;\textcolor{mygray}{\pm 0.36}$          & $84.33 \;\;\textcolor{mygray}{\pm 0.28}$          & $89.37 \;\;\textcolor{mygray}{\pm 0.47}$          & $92.44 \;\;\textcolor{mygray}{\pm 0.11}$          & $77.12\;\;\textcolor{mygray}{\pm 0.80}$ \\
            CAE                     & $83.10 \;\;\textcolor{mygray}{\pm 1.23}$          & $73.19 \;\;\textcolor{mygray}{\pm 0.82}$          & $75.82 \;\;\textcolor{mygray}{\pm 2.31}$          & $80.70 \;\;\textcolor{mygray}{\pm 2.93}$          & $82.72 \;\;\textcolor{mygray}{\pm 0.80}$          & $63.10 \;\;\textcolor{mygray}{\pm 6.51}$ \\
            \gls{gjsd}              & $84.38 \;\;\textcolor{mygray}{\pm 1.50}$          & $74.13 \;\;\textcolor{mygray}{\pm 0.64}$          & $77.56 \;\;\textcolor{mygray}{\pm 0.82}$          & $82.10 \;\;\textcolor{mygray}{\pm 3.72}$          & $84.78 \;\;\textcolor{mygray}{\pm 1.04}$          & $68.43 \;\;\textcolor{mygray}{\pm 7.75}$ \\
            IP-CAE                  & \bf{94.07}\;\;\textcolor{mygray}{$\pm$0.37}  &\bf{82.68}\;\;\textcolor{mygray}{$\pm$\bf{0.80}}   &\bf{91.85}\;\;\textcolor{mygray}{$\pm$\bf{0.55}}     &\bf{97.92}\;\;\textcolor{mygray}{$\pm$\bf{0.57}}   &\bf{93.71}\;\;\textcolor{mygray}{$\pm$\bf{0.62}}   & \bf{94.26} \;\;\textcolor{mygray}{$\pm$ \bf{1.48}} \\
            \end{tabular}
    \end{small}
\end{center}
\end{table*}

In Section~\ref{sec:experiments} we demonstrate state-of-the-art performance of IP-CAE compared to the methods listed above.


\paragraph{Other feature selection methods} Deep Lasso \citep{cherepanovaPerformanceDrivenBenchmarkFeature2023} is another generalization of the Lasso, different from LassoNet. Instead of penalizing the weights directly, the authors suggest penalizing the gradient and recovering the classic Lasso as a special case of their method. Another approach is training sparse neural networks where the sparse input layer naturally performs feature selection \cite{louizosLearningSparseNeural2018a, sokar_where_2022}. Although not trivially extended to neural networks, sparse priors formalize and extend the Lasso approach \citep{carvalhoHandlingSparsityHorseshoe2009, rockovaSpikeandSlabLASSO2018}. Deep Knockoffs \citep{romanoDeepKnockoffs2020} use deep generative models to enhance knockoff machines \citep{barberControllingFalseDiscovery2015}, a powerful method for statistical variable selection. Selecting measurements in compressed sensing, an important problem in medical imaging, has seen the development of similar methods to those in feature selection \citep{bakkerExperimentalDesignMRI2020a, huijbenDeepProbabilisticSubsampling2019}. Finally, instance-wise feature selection extends the problem of global feature selection to predict selections per sample, providing a hard, thresholded explanation of the network's prediction \citep{yoonINVASEInstancewiseVariable2018, chenLearningExplainInformationTheoretic2018}. 

\textbf{End-to-end learnable EEG channel selection} \citet{strypsteen2021endtoend} focus on a unified approach for selecting channels in electroencephalogram (EEG) recordings. They employ a concrete layer for feature selection and propose a heuristic regularization to penalize duplicate channel selections and show its effectiveness. We found that its performance requires careful tuning of three hyperparameters.

\textbf{Learning randomly perturbed structured predictors for direct loss minimization} \citet{indelmanLearningRandomlyPerturbed2021} propose learning the variance of the Gumbel noise perturbation in structured prediction. Although random perturbations are useful, they may mask the underlying signal during learning. By learning the variance of the perturbation noise, \citet{indelmanLearningRandomlyPerturbed2021} achieve superior performance to both fixed and zero noise settings.

\section{Experiments} \label{sec:experiments}



In this section, we evaluate our proposed method on several datasets. \Cref{tab:datasets} in \cref{app:experimental-details} provides an overview of the datasets used. Across all experiments, we perform an ablation of the standard \gls{cae}, only \gls{ip}, only the \gls{gjsd} term, and both \gls{ip} and the \gls{gjsd} term. The hyperparameter $P$ in $\Psi = \mathbb{R}^{K \times P}$ controlling the embedding dimension of the \gls{ip} selector layer, is not tuned in our experiments. Instead, we simply set $P=D$, where $D$ is the number of features for each dataset. The regularization strength hyperparameter for \gls{gjsd}, $\lambda$, was tuned in $\{0, 0.0005, 0.005, 0.05\}$. Following CAE, we use a fixed learning rate of 0.001 with the Adam optimizer with moving-average coefficients $\boldsymbol{\beta}=(0.9, 0.999)$ and no weight decay, for all experiments and datasets. 
We train every model for 200 epochs, and select the weights corresponding to the best validation loss for test set evaluation.
In all experiments, unless otherwise specified, we use an MLP with one hidden layer of 200 nodes for the decoder network. For the hidden activation, we use LeakyReLU with a slope of 0.2. 
For all experiments, we perform 10 repetitions and report the mean quantity. Any confidence intervals correspond to one standard deviation. 

\textbf{MNIST and MNIST-Fashion} \citep{mnist, fashion-mnist} consist of $28\times28$ greyscale images depicting digits and clothing items respectively. The supervised classification task is to predict the item from the pixel values.

\textbf{ISOLET} \citep{isolet} consists of preprocessed speech data of test subjects speaking all 52 letters of the English alphabet. The supervised classification task is to predict the spoken letter from the 617-dimensional speech data.

\textbf{COIL-20} \citep{coil20} consists of $32\times32$ greyscale images depicting 20 items, photographed on a rotating turntable at 5-degree increments (72 photos per item). The supervised classification task is to identify the item given the image.

\textbf{Smartphone Dataset for Human Activity Recognition} \citep{smartphone-har} consists of sensor data collected from 30 subjects performing 6 activities while wearing a smartphone. The classification task is to predict the action from the sensor signals.

\textbf{Mice Protein Expression} \citep{mice-protein-original-paper} consists of protein expressions from two groups of mice; control and trisomic mice. The supervised classification task is to predict the label consisting of the group, stimulation, and treatment of the mice. 7 proteins have missing values for one or more mice and these values were imputed with the average protein expression level of examples belonging to the same class of mice, following the dataset authors.

\textbf{Reconstruction Error}
We train end-to-end and report the reconstruction error as the normalized Frobenius norm $\norm{X - \hat X}_F / D$.

\textbf{Classification Accuracy}
We train end-to-end and report the classification in a supervised setting and report the top-1 accuracy.

\begin{figure}
    \centering
    \includegraphics[width=0.9\linewidth]{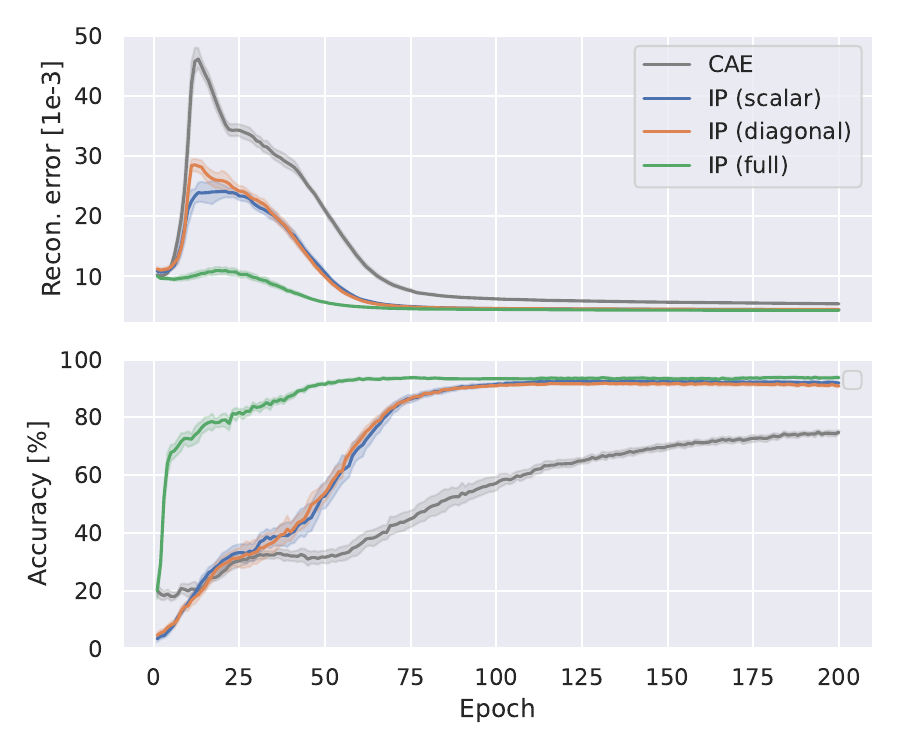}
    \caption{\textbf{IP parametrizations}. Validation results for CAE compared against three parametrizations of the linear IP weights on the ISOLET dataset.}
    \label{fig:ip-ablation}
\end{figure}

\subsection{Improved Training and Generalization with \gls{ip}}
\begin{figure*}[ht!]
    \centering
    \begin{subfigure}[b]{\textwidth}
         \centering
         \includegraphics[width=\textwidth]{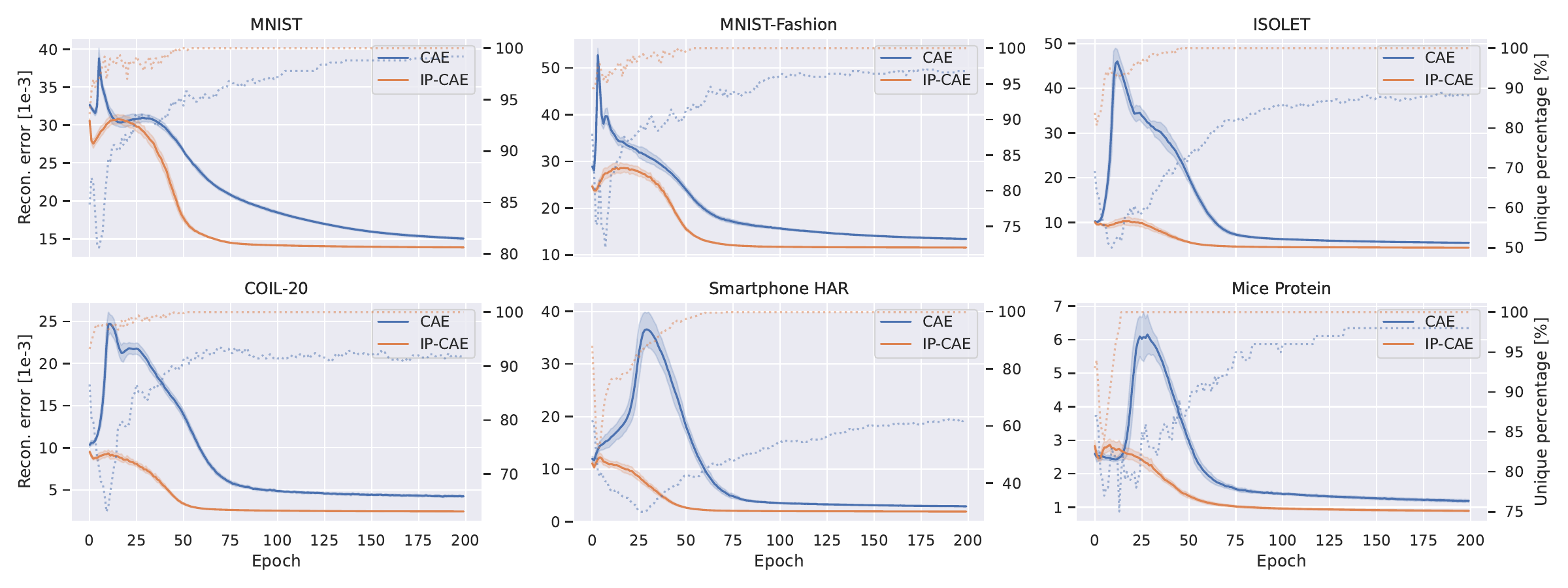}
         \caption{Reconstrucation error}
         \label{fig:recon-curves}
     \end{subfigure}
     \begin{subfigure}[b]{\textwidth}
         \centering
         \includegraphics[width=\textwidth]{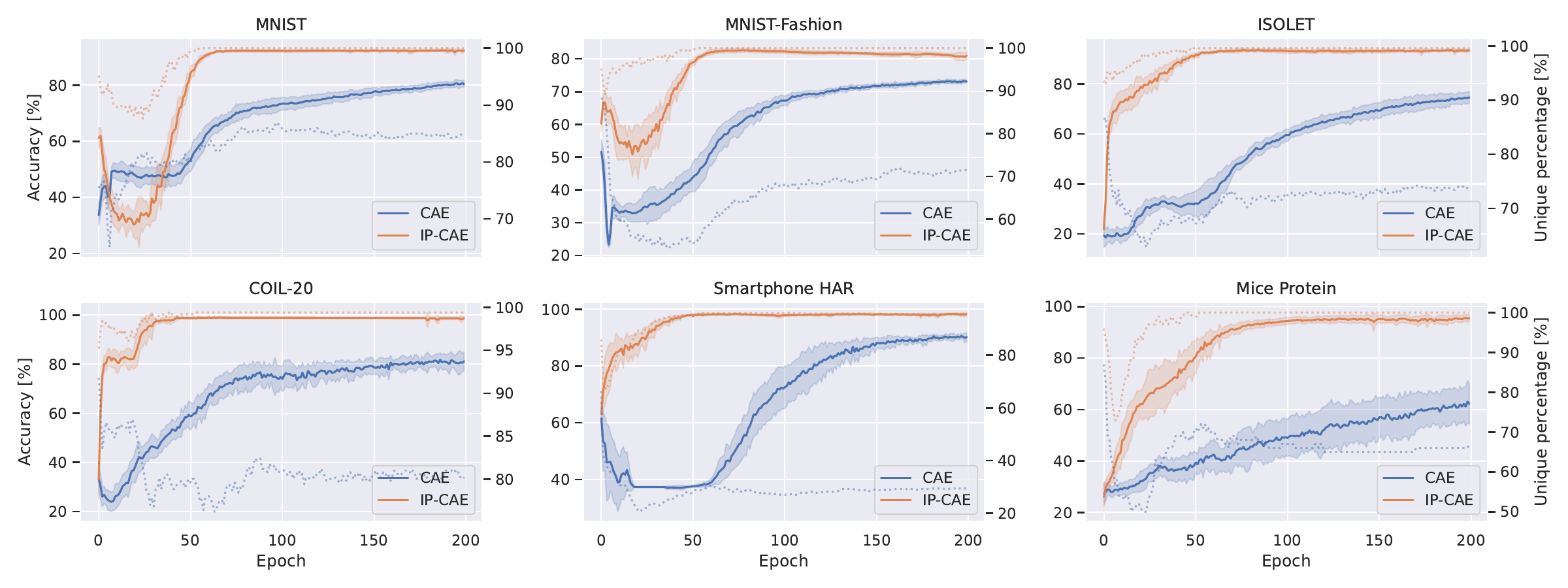}
         \caption{Accuracy}
         \label{fig:acc-curves}
     \end{subfigure}
     \caption{\textbf{Training Comparison}. Comparisons CAE and IP-CAE for (a) reconstruction error, (b) accuracy on the validation data throughout training. For IP-CAE, we let $P = D$. The mean unique percentages (definition \ref{def:unique-percentage}) is shown by the dotted lines.}
     \label{fig:training-curves}
\end{figure*}

\Cref{fig:training-curves} shows a significant improvement in training stability and convergence speed in both reconstruction and discriminative tasks. The curves represent our improved model (IP-CAE) with the original (\gls{cae}) across six common feature selection benchmarks. Each training was repeated for 10 random initializations, and the lines represent the mean validation loss (\Cref{fig:recon-curves}) and accuracy (\Cref{fig:acc-curves}), with one standard deviation indicated by the line width. Furthermore, IP-CAE converges to a lower validation error and higher accuracy than the original \gls{cae}. We also observe a significant speedup on all datasets (\cref{tab:speedup}).

\begin{figure}
    \centering
    \includegraphics[width=0.9\linewidth]{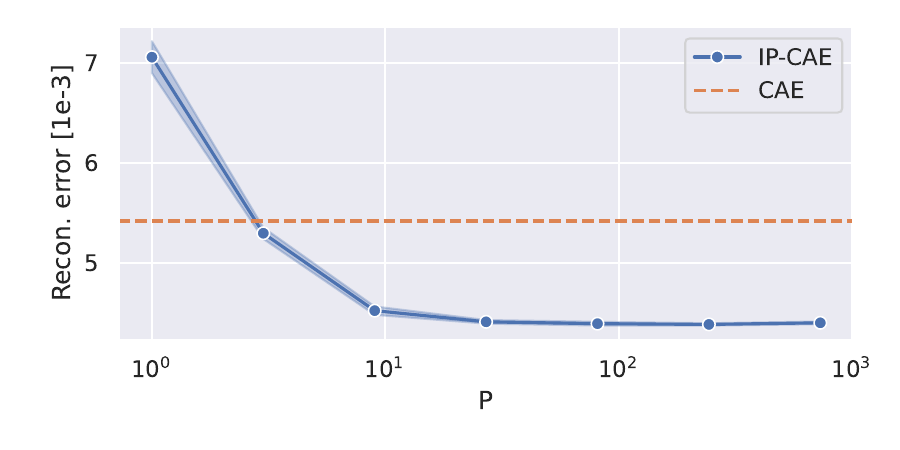}
    \caption{\textbf{Varying P}. Test set performance on ISOLET for varying size of \gls{ip} $P$. The mean reconstruction error with CAE is included as a horizontal line.}
    \label{fig:varying-P}
\end{figure}

While we do not tune P, we include an ablation of the setting of P for one dataset, Isolet, in \Cref{fig:varying-P}. We conclude that the \gls{ip} is not sensitive to a specific setting of P so long as it is sufficiently large, \textit{i.e.} $P \approx D$. We find that the bias term in \gls{ip} (\Cref{eq:ip}) is redundant and does not affect performance.

To verify that this effect applies in general and not just for a specific setting of $K$, we vary $K$ in $\{25, 50, 75, 100, 125\}$ on the ISOLET dataset. \Cref{fig:varying-k} in \cref{app:additional-experiments} confirms that the results are valid regardless of $K$.

As a lower bound on performance, we include a comparison with the proposed GJSD regularization method which explicitly encourages unique selections. We find that such explicit encouragement outperforms CAE, but is not as effective as IP.


\begin{table}
    \centering
    \caption{\textbf{Speedup}. The mean speedup of IP-CAE compared to CAE, in terms of IP-CAE surpassing the performance of CAE (on validation data) trained for 200 epochs.}
    \begin{center}
    \begin{small}
        \begin{sc}
            \begin{tabular}{{p{3cm}}*{2}{p{2cm}}}
            Dataset         & Recon. error & Accuracy \\
            \hline
            MNIST           & 3.00$\times$ & 4.00$\times$\\
            MNIST-Fashion   & 3.38$\times$ & 4.60$\times$\\
            ISOLET          & 3.83$\times$ & 18.77$\times$\\
            COIL-20         & 4.15$\times$ & 25.68$\times$\\
            Smartphone HAR  & 4.15$\times$ & 3.70$\times$\\
            Mice Protein    & 2.53$\times$ & 12.92$\times$ \\
            \end{tabular}
        \end{sc}
    \end{small}
\end{center}
    \label{tab:speedup}
\end{table}

\begin{figure*}[ht]
    \begin{subfigure}[b]{0.5\linewidth}
        \centering
        \includegraphics[width=\linewidth]{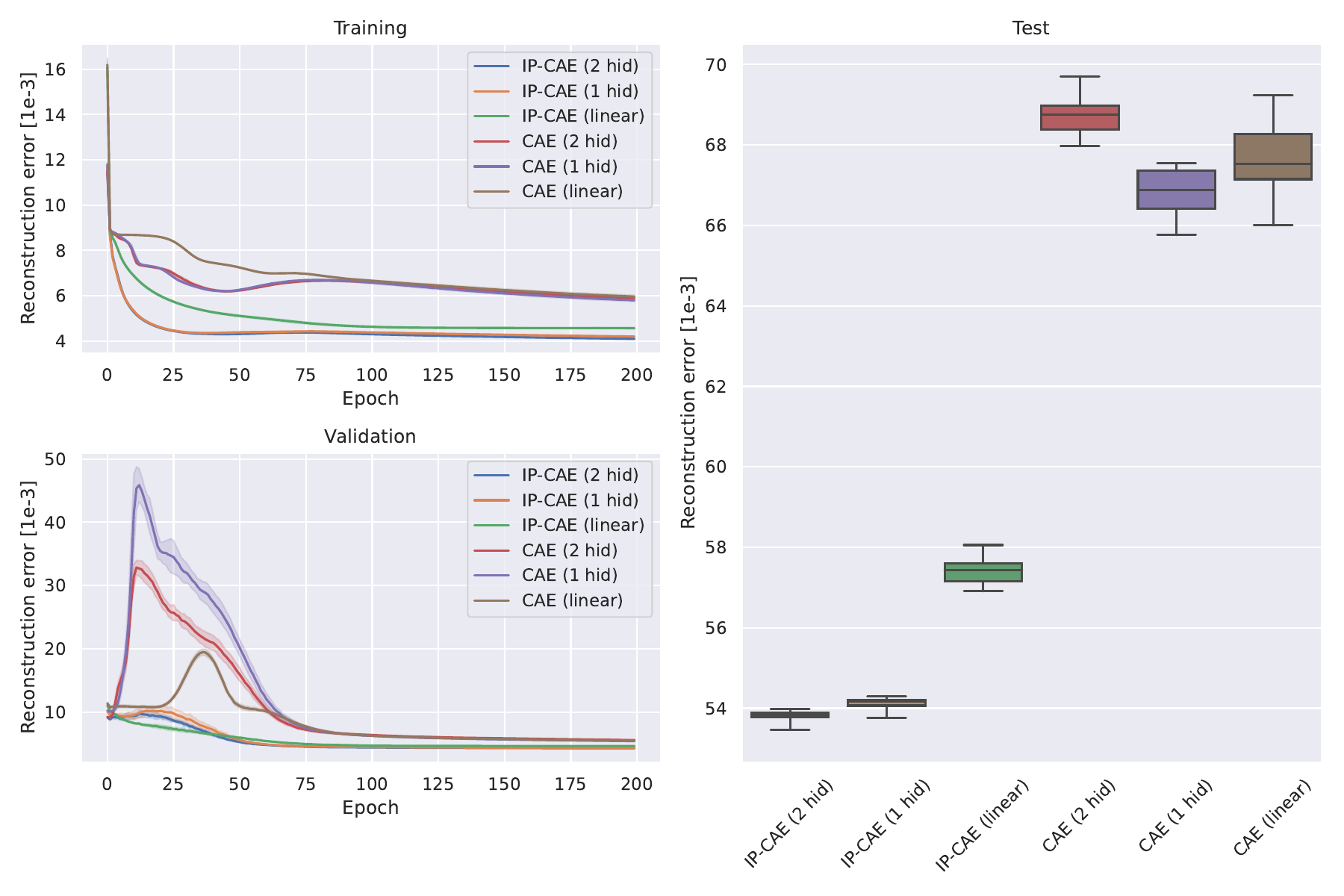}
        \caption{Reconstruction}
    \end{subfigure}
    \begin{subfigure}[b]{0.5\linewidth}
        \centering
        \includegraphics[width=\linewidth]{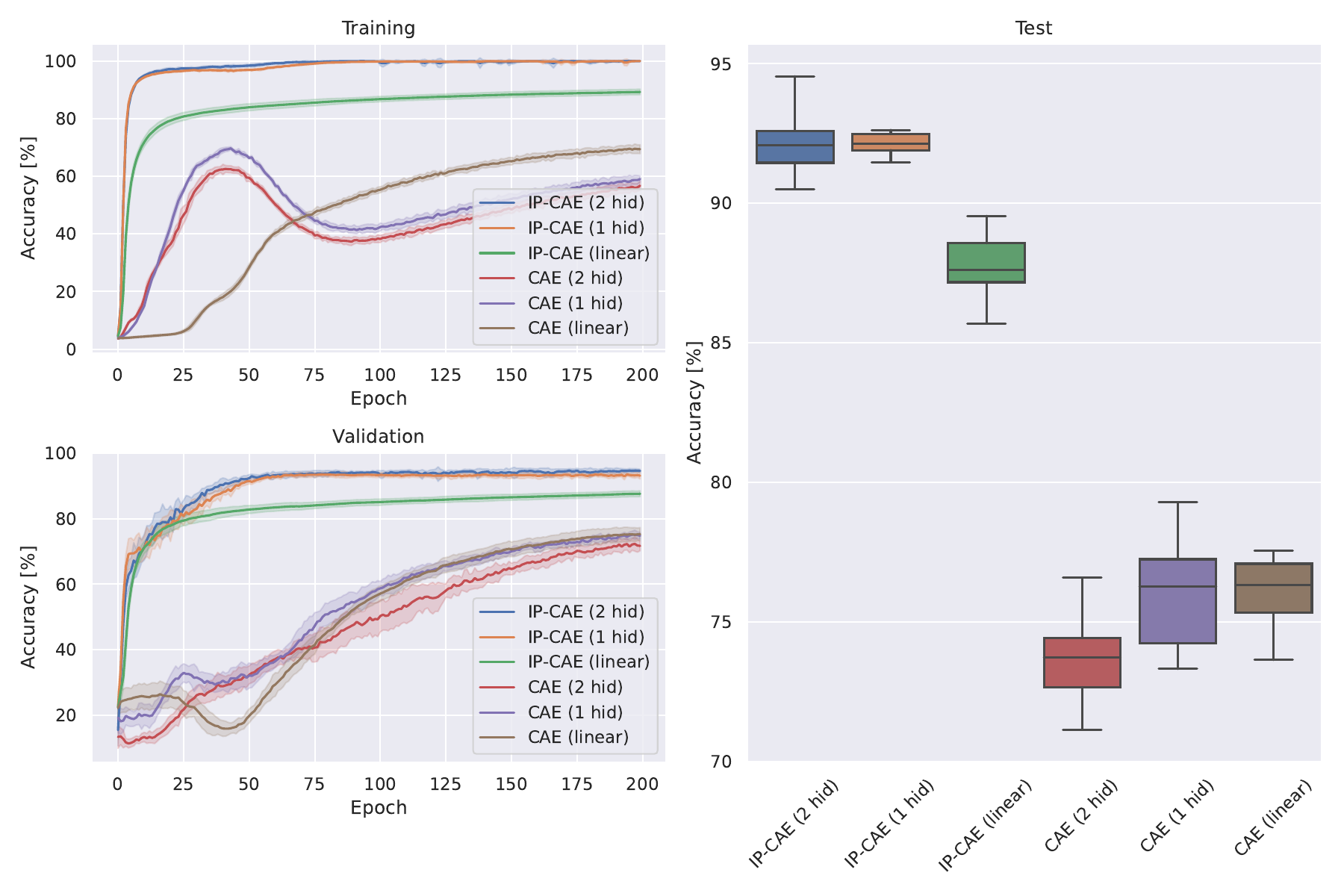}
        \caption{Classification}
    \end{subfigure}
    \caption{\textbf{Hidden Layers}. Results for varying decoder architectures on the ISOLET dataset, with and without \gls{ip}. Three architectures are considered; linear, one hidden layer with 200 nodes, and two hidden layers with 200 nodes each. Unlike the original CAE, IP-CAE benefits significantly from additional decoder capacity. Boxes are quartiles and whiskers are min-max.}
    \label{fig:hidden}
\end{figure*}

\subsection{Special Cases of IP-CAE}
As addressed in \Cref{sec:method:update_rule}, \gls{cae} is a special case of IP-CAE with $\boldsymbol{W} = \mathbf{I}$ and $P=D$. Two additional special cases of IP-CAE that preserve the learning rate scaling properties (\Cref{sec:appendix:update_rules}) in the update rule while offering even simpler formulations are as follows:

\textbf{Single Scalar}. The first alternative formulates $\boldsymbol{W}$ as $\boldsymbol{W}=w\boldsymbol{I}$, where $w$ is a single learnable scalar parameter. This approach simplifies the complexity of $\boldsymbol{W}$ to a single degree of freedom.

\textbf{Diagonal Matrix}. The second alternative represents $\boldsymbol{W}$ as $\boldsymbol{W}=\text{diag}(\mathbf{w})$, with $\mathbf{w}$ being a vector of learnable parameters.

Our empirical analysis, presented in \Cref{fig:ip-ablation}, contrasts these two simplified forms of $\boldsymbol{W}$ with the full matrix version and the standard CAE configuration.
Both the scalar and diagonal versions demonstrate enhancements over CAE in stability and final performance. This improvement underscores the significance of the learning rate scaling property, as maintained in these simpler forms.

Most notably, allowing $\boldsymbol{W}$ to be a full matrix results in the most pronounced improvements in terms of training efficiency. This observation strongly suggests that the 
$\boldsymbol{W}\boldsymbol{W}^T$ term in the gradient transformation (\Cref{eq:ip_update_rule}) plays a significant role in the model's ability to learn complex embeddings to represent features.

\subsection{Hidden Layers}
It is worth noting that the main promise of neural network-based embedded feature selection is that it can be non-linear. We observe that for the original \gls{cae}, training is unstable with spikes in validation error, hindering a smooth convergence to the optimal solution. The problem worsens for decoders with multiple hidden layers. This is illustrated in \cref{fig:hidden}, where we investigate this issue for a linear decoder, an MLP decoder with one hidden layer, and an MLP decoder with two hidden layers.
Our improved parameterization results in a smooth descent into a final validation loss that is lower than for the original CAE regardless of decoder complexity, and thus allows for the joint optimization of feature selection and non-linear decoders.

\subsection{Comparison with STG and LassoNet}
For completeness, we compare the test performance of the vanilla and IP-CAE with other baselines such as STG and LassoNet, in both reconstruction and classification settings in \cref{tab:reconstruction,tab:classification}. Unlike CAEs, both STG and LassoNet cannot optimize for a specific number of selected features directly, instead, they require the hyperparameter $\lambda$, which denotes the strength of regularization, to be specified, which in turn affects the number of selected features. LassoNet also requires an additional hyperparameter $M$, which denotes the hierarchy coefficient, however, following \citep{lemhadri_lassonet_2021} we use $M=10$ for all the datasets. We run extensive ablations over the hyperparameter $\lambda$ and choose the value that returns 50 feature selections for all our datasets (with the exception of MICE, where we have 10 feature selections). For STG and LassoNet, we use a single hidden layer MLP with 200 dimensions and ReLU activation during feature selection. Later, we retrain a single hidden layer MLP with 200 dimensions and ReLU from scratch to report the test accuracy, for each dataset.

We used the official code repositories of STG and LassoNet for all our experiments. For STG, during and after feature selection, the networks were trained for 100 epochs for all datasets. For LassoNet, during feature selection, we used the default setting from the official repository to train the initial dense network for 1000 epochs followed by 100 epochs of training for each sparse network in the iterative process (increasing $\lambda$). After feature selection, we again run 100 epochs of network training with the corresponding selected features for all datasets.

For both CAE and IP-CAE, we refrain from retraining the decoder, but instead directly evaluate it using the jointly learned decoder. We chose this approach because it aligns with the core principle of embedded feature selection, which is to utilize features non-linearly and jointly optimize for feature selection and the non-linear training objective.

We demonstrate that CAE falls behind STG and LassoNet by a significant margin both for reconstruction and classification. But IP-CAE significantly outperforms them in test accuracy (\cref{tab:classification}) and reconstruction error (\cref{tab:reconstruction}) on all datasets. Additionally, being a stochastic method, CAE is prone to high variance. This problem is reduced slightly in IP-CAE on most datasets, as shown in \cref{fig:training-curves}.
\section{Discussion} \label{sec:discussion}
In this paper, we addressed the practical challenges of training CAEs. We proposed IP-CAEs which implicitly alleviate redundant features and instability. Our approach achieves state-of-the-art reconstruction error and accuracy for all datasets considered, up to 20 times faster than vanilla CAE. 

While, in this paper, we establish the empirical effectiveness of IP-CAE across a wide range of datasets and tasks, and argue for it from the lens of implicit overparametrization, the remarkable results motivates the need for a more formal study as a future direction.

With the goal of understanding IP-CAE's success, we introduced \gls{gjsd} regularization, which forces unique selections. This baseline significantly improves CAE in every dataset and task, but falls behind \gls{ip}, as well as LassoNet and STG. This, interestingly, indicates that the effect of IP cannot be solely attributed to removing duplicate features. \gls{ip} and \gls{gjsd} are not exclusive and can be combined, but we found that adding \gls{gjsd} regularization to IP-CAE not to improve results significantly.

Finally, the \gls{ip} method we present is, in principle, generalizable to Gumbel-Softmax distributions beyond feature selection which is left for future work.

\section*{Impact Statement}
This paper presents work whose goal is to advance the field of Machine Learning. There are many potential societal consequences of our work, none of which we feel must be specifically highlighted here.

\section*{Acknowledgements}
We thank the anonymous reviewers for their valuable feedback and suggestions. This work was partially supported by KTH Digital Futures, the   Swedish eScience Research Centre (SeRC), the Swedish Foundation for
Strategic Research grants BD15-0043 and ID19-0052, the Marie Skłodowska-Curie Actions project "MODELAIR" through grant no. 101072559, and Wallenberg AI, Autonomous Systems and Software Program (WASP) funded by the Knut and Alice Wallenberg Foundation. The computations were enabled by Alvis cluster access provided by the National Academic Infrastructure for Supercomputing in Sweden (NAISS) at Chalmers University of Technology partially funded by the Swedish Research Council through grant agreement no. 2022-06725 as well as the Berzelius resource provided by the Knut and Alice Wallenberg Foundation at the National Supercomputer Centre.

\bibliography{references}
\bibliographystyle{icml2024}

\newpage
\appendix
\clearpage
\section{Experimental Details} \label{app:experimental-details}
In this appendix, we provide a detailed description of the experimental setup.

\subsection{Datasets}
\Cref{tab:datasets} provides an overview of all datasets used in experiments, the type of data, number of samples $N$, features $D$, selected features $K$, and classes $C$.
\begin{table*}
    \caption{\textbf{Datasets}. An overview of the datasets used. The number of samples $N$ includes both training and test data. For MNIST and MNIST-Fashion, the data used is a random subset of the full data.}
    \label{tab:datasets}
    \begin{small}
    \begin{sc}
        \begin{center}
            \begin{tabular}{l|c|c|c|c|c}
                Name & Input type &Samples ($N$) & Features ($D$) & Selected ($K$) & Classes ($C$) \\
                \hline
                MNIST & Grayscale image & 10500 & 784 & 50 & 10 \\
                MNIST-Fashion\ & Grayscale image & 10500 & 784 & 50 & 10 \\
                ISOLET & Speech & 7797 & 617 & 50 & 26 \\
                COIL-20 & Grayscale image & 1440 & 1024 & 50 & 20\\
                Smartphone HAR & Sensor time series & 10299 & 561 & 50 & 6 \\
                Mice Protein & Protein expression & 1080 & 77 & 10 & 8
            \end{tabular}
        \end{center}
    \end{sc}
\end{small}
\end{table*}

\subsection{Clarification of hyperparameters}
We use the same temperature annealing schedule as CAE, and the same maximum temperature $T_{0}=10$ and minimum temperature $T_{B}=0.01$. We searched the regularization strength hyperparameter of the GJSD term in \{5.00E-04, 5.00E-03, 5.00E-02\} for both reconstruction and classification. The optimal settings found are listed in \cref{tab:lambda-table}.

\begin{table}
    \begin{small}
    \begin{sc}
    \centering
    \caption{\textbf{GJSD settings}. Optimal settings of the GJSD regularization strength hyperparameter $\lambda$ using the original CAE parametrization.}
    \begin{tabular}{c|c|c}
         Dataset&  Classification& Reconstruction\\
         \hline
         MNIST&  5.00E-02& 5.00E-02\\
         MNIST-Fashion&  5.00E-02& 5.00E-02\\
         Smartphone HAR&  5.00E-02& 5.00E-03\\
         COIL-20&  5.00E-02& 5.00E-02\\
         ISOLET&  5.00E-02& 5.00E-03\\
         Mice Protein&  5.00E-02& 5.00E-03\\
    \end{tabular}
    \label{tab:lambda-table}
    \end{sc}
    \end{small}
\end{table}

\subsection{Code}
The source code is available at \url{https://github.com/Alfred-N/IP-CAE}.

We have taken considerable care to ensure the ease of reproducibility of all our results. For each dataset, we have included a configuration file named \lstinline{<dataset>/base.yaml} which contains the necessary hyperparameters to run CAE exactly as we did in our paper for the reconstruction task. To run with our proposed IP, simply specify the \lstinline{--dim_ip} optional argument when executing our training script, which refers to the dimensionality of the IP vectors, namely $P$. Example: \lstinline{python src/main_pl.py --config=configs/ISOLET/base.yaml --dim_ip=617}. The setting of \lstinline{dim_ip} ($P$) used throughout all our experiments with IP was configured to match the feature dimensions of each dataset, $D$, which can be found in Table 1 of the main report. Similarly, the corresponding configs for the classification task can be found as \lstinline{<dataset>/classification.yaml}.

We log training metrics with WandB. To start tracking without a WandB account, simply run \lstinline{src/main_pl.py} and select option \lstinline{(1) Private W&B dashboard, no account required} when prompted. Note that this requires an internet connection. To run offline, select option \lstinline{(4) Don't visualize my results}.

To further facilitate reproducibility, we include a script, \lstinline{src/fs_datasets.py}, for downloading all datasets used in this paper, which includes functions that return the exact train/test/validation splits that were used. The data will be automatically downloaded into \lstinline{--data_root_dir} when running \lstinline{src/main_pl.py}.

For all of our experiments that we repeated for 10 seeds, we used fixed seeds \{11, 22, 33, 44, 55, 66, 77, 88, 99, 1010\}. Thus, all results can be reproduced exactly and deterministically if specifying the (integer) argument \lstinline{--seed}.

Finally, \lstinline{--IP_weights} flag can be used to specify different weight options of IP such as \lstinline{scalar}, \lstinline{diag} or \lstinline{shared}. Note that the general version of IP we describe in the main paper refers to the \lstinline{shared} option.

\subsection{Data and preprocessing}
For COIL-20 we use the version of the dataset provided by \cite{li_feature_2017}. For MNIST and Fashion-MNIST, we use the versions provided in Torchvision. For the other datasets, ISOLET, Smartphone HAR, and Mice protein we use the version provided at UCI \citep{isolet, smartphone-har, mice-protein-original-paper}.

We identified a potential bug in the preprocessing of the Mice Protein dataset used by CAE. They impute missing values with a "filling value" of $-10^{5}$ and then take column averages of each protein expression and replace the filling value. The expression levels are generally in the order of magnitude of $10^0$ to $10^1$, which means the average is dominated by the filling value rather than the signal. Additionally, they overwrite the same array they use to calculate averages on the fly instead of inputting the data in a new array.

We instead use the imputation method described by the authors of the Mice Protein dataset \citep{mice-protein-original-paper}, which means averaging missing protein expression values with the average expression corresponding to that protein for the same class of mice.

Additionally, CAE computes their min-max scaling based on the statistics of the full dataset. We instead calculate the min-max scaling statistics only on the training split and then use them to scale the validation and test split accordingly.

\subsection{Compute infrastructure}
We used an external cluster with T4 and A40 GPUs. Each model was trained on a single GPU.

\section{Additional Experiments} \label{app:additional-experiments}
In this section, we provide additional experiments that were left out of the paper due to the space limit.

\subsection{Extended training}
We compare the convergence with an increased number of epochs in the ISOLET dataset. We increase the number of epochs to 1000 (from 200 in our other experiments).
This way, the annealing schedule of the temperature is stretched over a longer period, which means a longer exploration phase with high randomness.

As mentioned by the CAE authors: \textit{"if the temperature is held low, the concrete selector layer is not able to explore different combinations of features and converges to a poor local minimum"}.

We find that this longer training drastically improves CAE, which converges to higher accuracy, lower reconstruction error and higher unique percentage, see \cref{tab:extended-training}. However, IP-CAE trained for 200 epochs still outperforms CAE trained for 1000 epochs. We emphasize that this is for illustrative purposes. Training for five times as many epochs is not an efficient solution to CAEs' undesirable training behavior. Interestingly, we observe that CAE does not achieve 100\% unique selections for classification on the ISOLET dataset, which seems to limit the resulting accuracy.

\begin{table}
    \begin{small}
    \begin{sc}
    \centering
    \caption{\textbf{Extended Training}. The mean test set performance on ISOLET for CAE and IP-CAE trained for 200 and 1000 epochs. The mean is computed using ten repetitions.}
    \begin{tabular}{l|c|c|c}
    Model & Epochs & Recon. error & Accuracy\\
    \hline
    CAE & 200 & 0.067 & 75.8 \\
    CAE & 1000 & 0.054 & 91.0 \\
    IP-CAE & 200 & 0.054 & \bf{91.9} \\
    IP-CAE & 1000 & \bf{0.053} & 91.4 \\
    \end{tabular}
    \label{tab:extended-training}
    \end{sc}
    \end{small}
\end{table}

\subsection{Number of selected features}
\begin{figure*}
    \centering
    \begin{subfigure}[b]{0.45\linewidth}
        \centering
        \includegraphics[width=\linewidth]{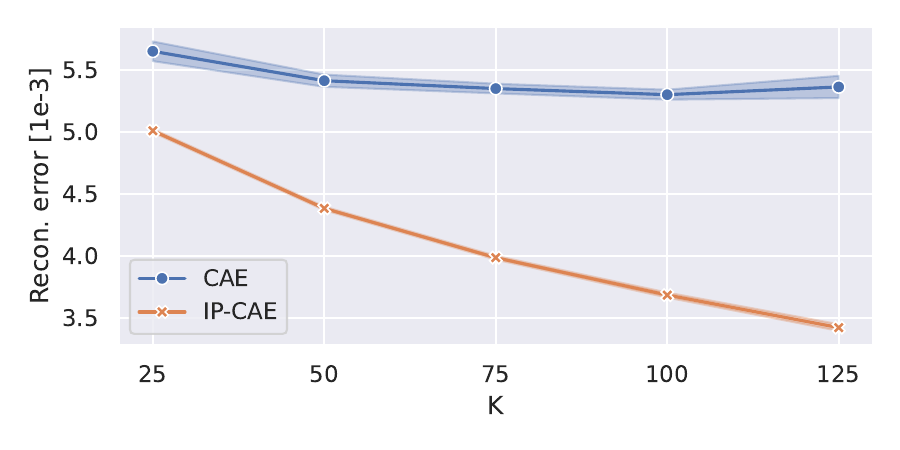}
        \caption{Reconstruction}
        \label{fig:varying-k-rec}
    \end{subfigure}
    \begin{subfigure}[b]{0.45\linewidth}
        \centering
        \includegraphics[width=\linewidth]{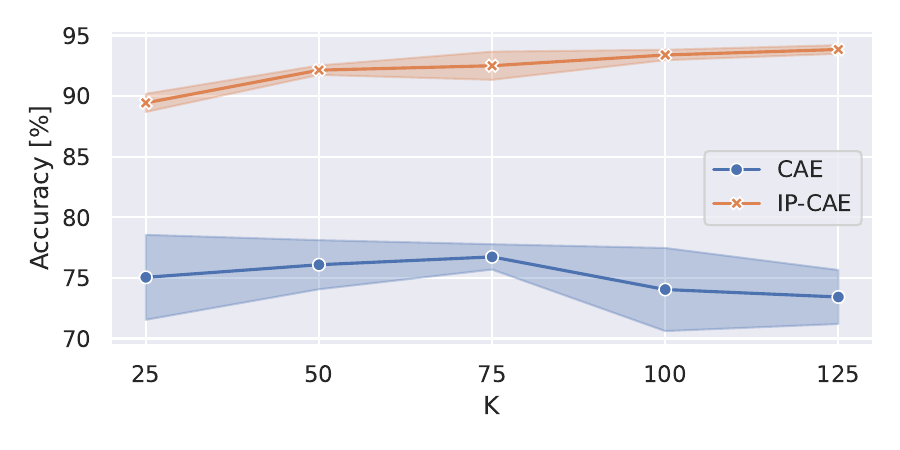}
        \caption{Classification}
        \label{fig:varying-k-acc}
    \end{subfigure}
    \caption{\textbf{Varying K}. Test set performance for (a) reconstruction and (b) classification while varying the number of features selected $K$ with and without \gls{ip} on ISOLET.}
    \label{fig:varying-k}
\end{figure*}

\subsection{Embedding dimensionality}
Here, we provide additional experiments showcasing the effect of the parameter $P$ on the ISOLET dataset.
\begin{figure*}
    \centering
    \begin{subfigure}[b]{0.329\textwidth}
        \centering
        \includegraphics[width=\textwidth]{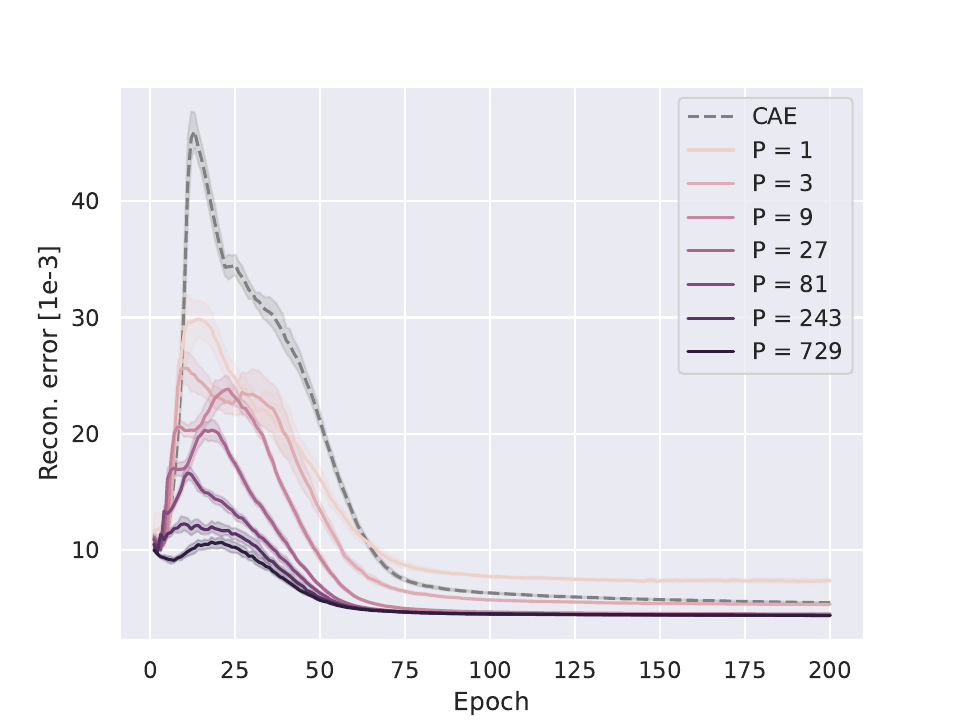}
        \caption{Reconstruction error}
    \end{subfigure}
    \begin{subfigure}[b]{0.329\textwidth}
        \centering
        \includegraphics[width=\textwidth]{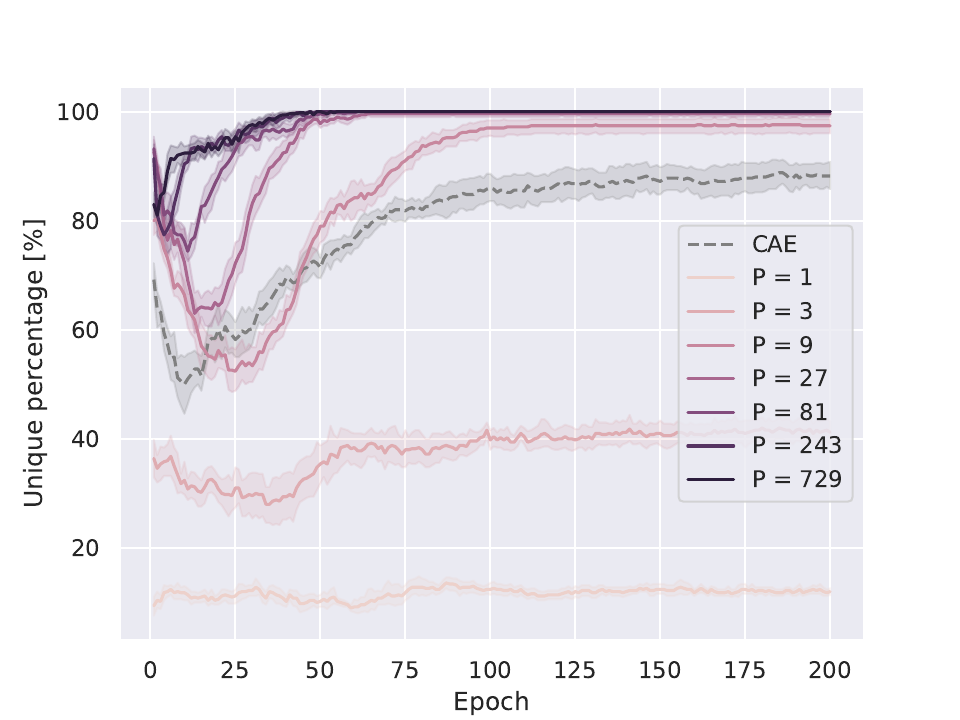}
        \caption{Unique percentage}
    \end{subfigure}
    \begin{subfigure}[b]{0.329\textwidth}
        \centering
        \includegraphics[width=\textwidth]{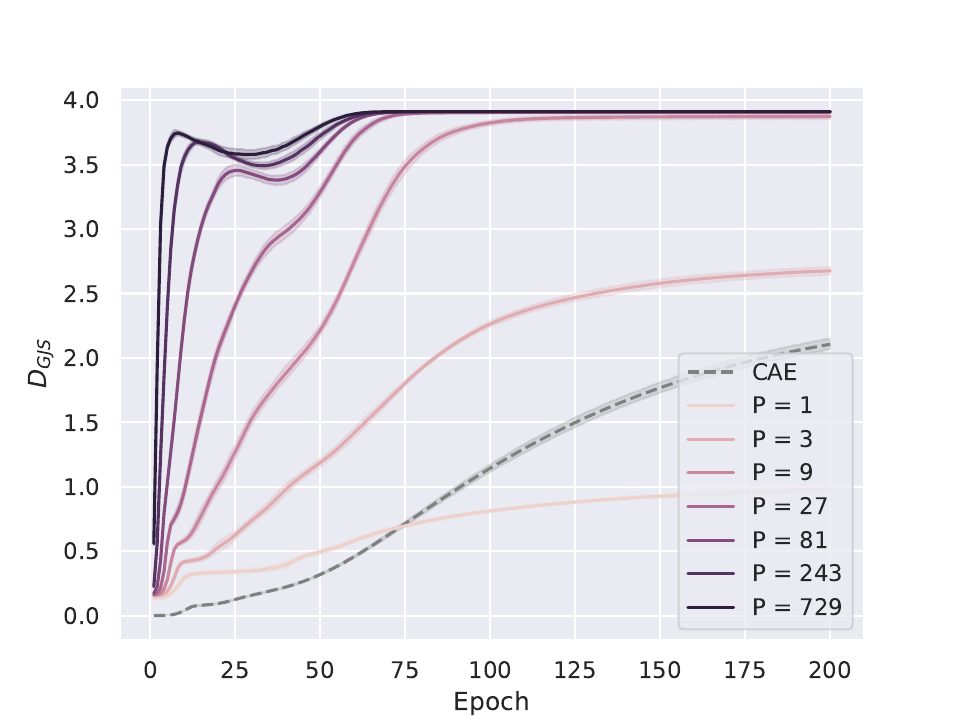}
        \caption{$D_{GJS}$}
    \end{subfigure}
    \caption{\textbf{Convergence with varying P}. Validation results for a) reconstruction error, b) unique percentage, and c) generalized Jensen-Shannon divergence. Plots are the mean of ten repetitions and confidence bounds show one standard deviation.}
    \label{fig:vary-p}
\end{figure*}
As evident from the results, IP-CAE outperforms CAE even when using a smaller number of parameters. We call this setting \textit{underparameterized}. Naturally, as the number of parameters decreases further, the IP-CAE performs worse than the CAE at a certain point. The training seems to improve with $P$, but not necessarily the end result if trained to convergence.


\subsection{Learning rate warmup}
\begin{figure*}
    \centering
    \begin{subfigure}[b]{0.49\textwidth}
        \centering
        \includegraphics[width=0.7\linewidth]{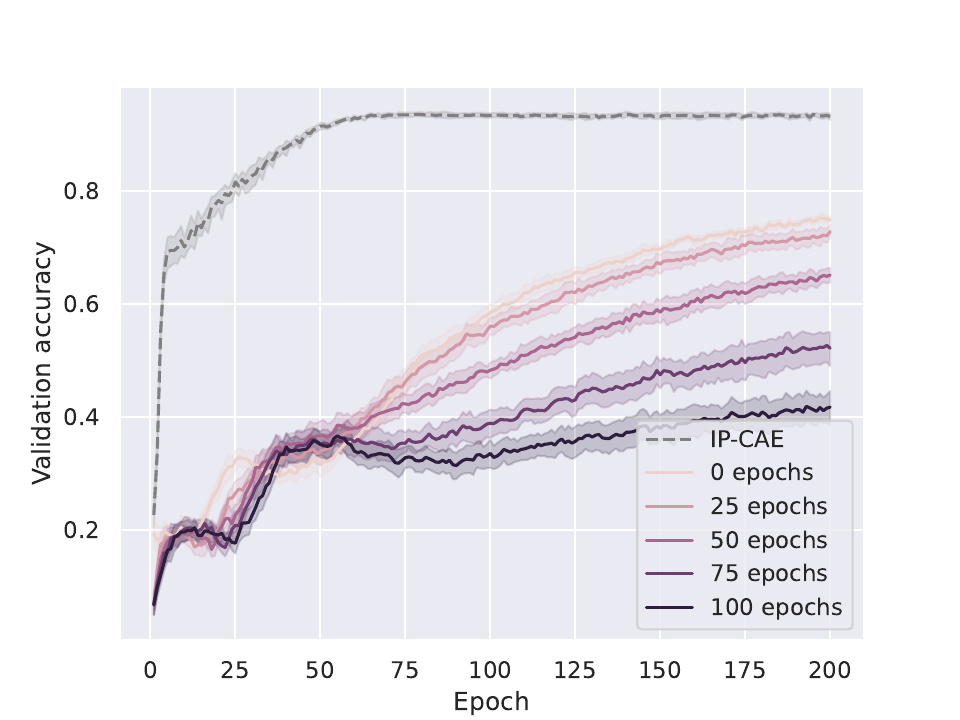}
        \caption{Learning rate warmup}
        \label{fig:warmup}
    \end{subfigure}
    \begin{subfigure}[b]{0.49\textwidth}
        \centering
        \includegraphics[width=0.7\linewidth]{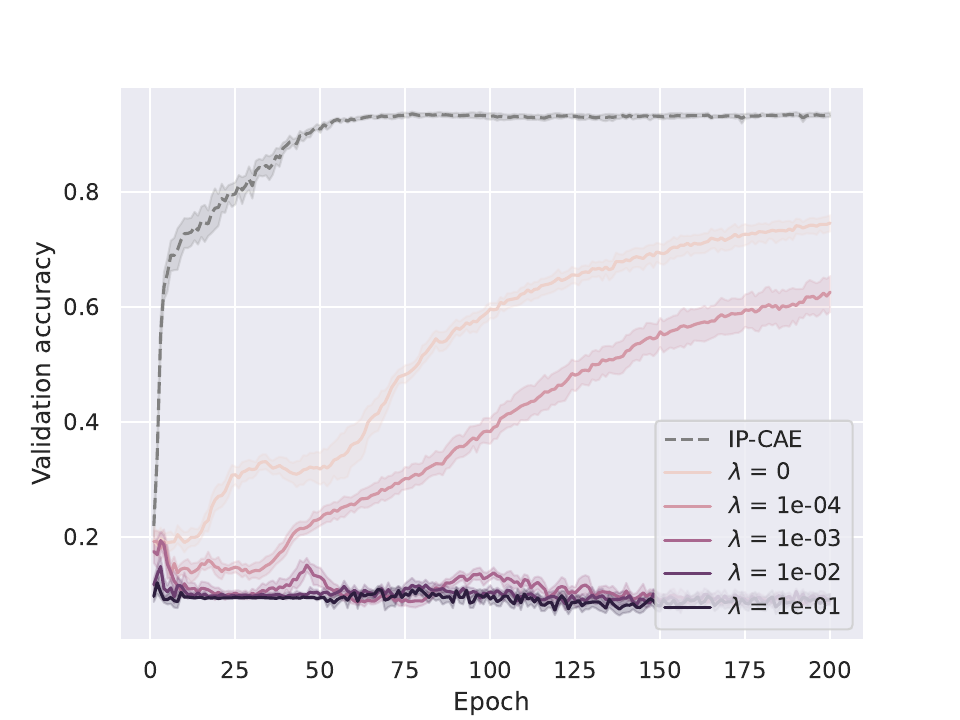}
        \caption{Weight decay}
        \label{fig:weight-decay}
    \end{subfigure}
    \caption{\textbf{Conventional training modifications}. An interesting question is whether the problems discussed in this work can be alleviated using conventional techniques like scheduling the learning rate (a) or through standard weight decay regularization (b). }
\end{figure*}

We use a warmup phase with a linearly increasing learning rate from $10^{-6}$ to $10^{-3}$ for the first $\{25, 50, 75, 100\}$ epochs out of 200. This reduces the spike in validation loss but leads to worse results (\cref{fig:warmup}). Our interpretation is that the lower learning rate causes the model to learn less during the critical early exploration phase, a phase that is critical to finding optimal minima \cite{balin_concrete_2019}.

\subsection{Weight decay}
We consider weight decay with parameters $\{10^{-4}, 10^{-3}, 10^{-2}, 10^{-1}\}$, all of which yield worse results (\cref{fig:weight-decay}). This is consistent with Bora et al. (2019), where no weight decay is used.

\subsection{Runtime}
We report the average run time on a T40 GPU in \cref{tab:wall_time}.
\begin{table}
    \begin{small}
    \begin{sc}
    \centering
    \caption{\textbf{Runtime}. The average time in minutes of training for 200 epochs, with and without IP.}
    \begin{tabular}{c|c|c|c|c}
         Dataset & \multicolumn{2}{c}{Classification} & \multicolumn{2}{c}{Reconstr.}\\
                        & CAE    & IP-CAE & CAE    & IP-CAE \\
         \hline
         MNIST          & 8.55   & 8.73   & 8.74   & 8.86   \\
         MNIST-Fashion  & 8.67   & 8.82   & 8.71   & 8.87   \\
         Smart. HAR & 5.47   & 5.71   & 5.07   & 5.15   \\
         COIL-20        & 5.28   & 5.67   & 4.95   & 5.40   \\
         ISOLET         & 6.10   & 6.21   & 5.85   & 5.94   \\
         Mice Protein   & 4.14   & 4.41   & 3.55   & 3.63   \\
    \end{tabular} 
    \label{tab:wall_time}
    \end{sc}
    \end{small}
\end{table}

\clearpage
\onecolumn
\section{IP-CAE Updates} \label{app:gradients}
\begin{figure*}[ht!]
    \centering
    \begin{subfigure}{0.329\linewidth}
        \centering
        \includegraphics[width=\linewidth]{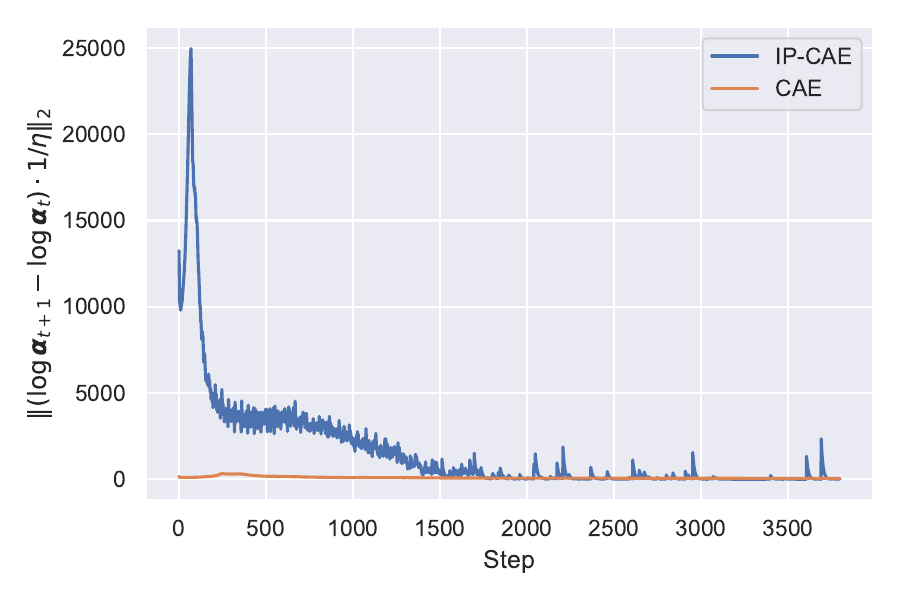}
        \caption{$\boldsymbol\alpha$}
    \end{subfigure}
    \begin{subfigure}{0.329\linewidth}
        \centering
        \includegraphics[width=\linewidth]{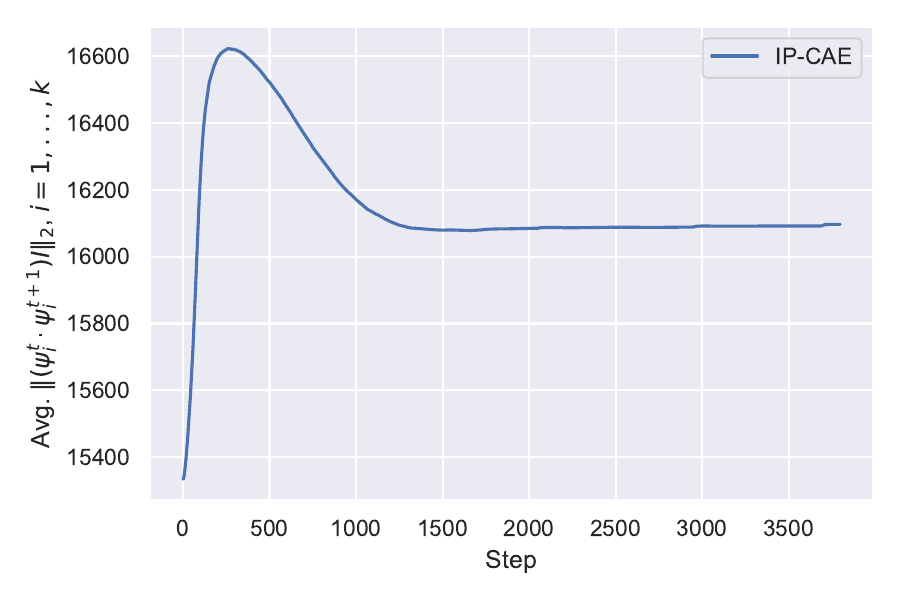}
        \caption{$\psi$}
    \end{subfigure}
    \begin{subfigure}{0.329\linewidth}
        \centering
        \includegraphics[width=\linewidth]{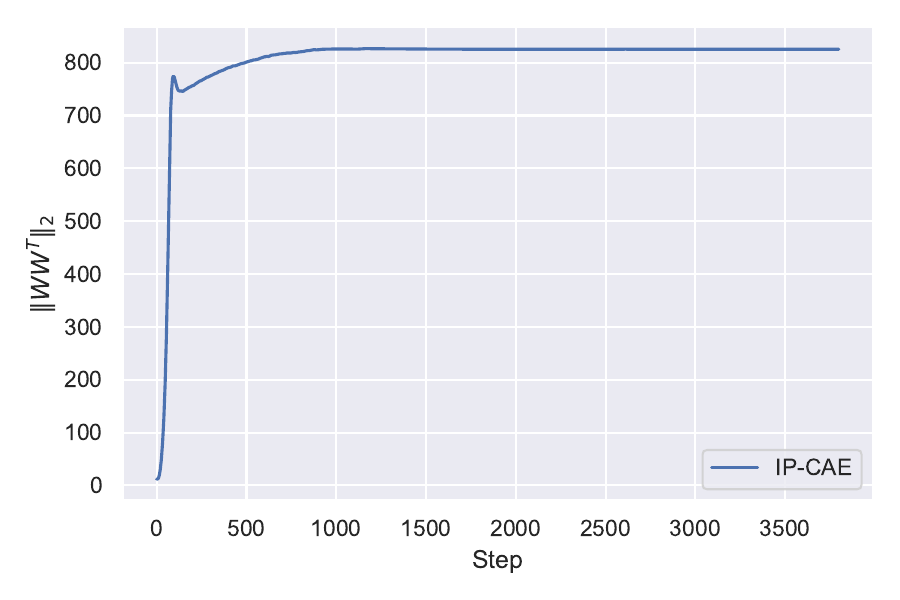}
        \caption{$\boldsymbol{W}$}
    \end{subfigure}
    \caption{\textbf{IP-CAE Update}. Components of the update rule related to (a) $\boldsymbol\alpha$, (b) $\psi$, and (c) $\boldsymbol{W}$ for each step throughout training for classification on ISOLET. In (a), the value is compared to its equivalent in the vanilla CAE.}
    \label{fig:updateplots}
\end{figure*}

In this appendix, we include additional plots showing how components of the IP-CAE update change over time (\Cref{fig:updateplots}).

\section{Update Rules for IP-CAE}
\label{sec:appendix:update_rules}
In this appendix, we derive the update rules for IP-CAE.

\textbf{Full Weight Matrix.} With a full $\boldsymbol{W}$ matrix, we have $\log \boldsymbol{\alpha}_i^{(t+1)} = \boldsymbol{W}^{(t+1)}\boldsymbol{\psi}_i^{(t+1)}$ and:
\begin{align}
        \boldsymbol{W}^{(t+1)} \leftarrow \boldsymbol{W} - \eta \nabla_{\boldsymbol{W}}\mathcal{L} \\
        \boldsymbol{\psi}_i^{(t+1)} \leftarrow \boldsymbol{\psi}_i - \eta \nabla_{\boldsymbol{\psi}_i}\mathcal{L}
\end{align}
Thus, to derive the update rule for $\log \boldsymbol{\alpha}_i^{(t+1)}$ we need to first derive $\nabla_{\boldsymbol{W}}\mathcal{L}$ and $\nabla_{\boldsymbol{\psi}_i}\mathcal{L}$, which are the gradients of the loss with respect to $\boldsymbol{W}$ and $\boldsymbol{\psi}_i$, respectively. We have:
\begin{align}
    \nabla_{\boldsymbol{W}}\mathcal{L} &= 
    \begin{bmatrix}
    \nabla_{\boldsymbol{W}_{1,1}}\mathcal{L} & \dots & \nabla_{\boldsymbol{W}_{1,D}}\mathcal{L} \\
    \vdots & & \\
    \nabla_{\boldsymbol{W}_{D,1}}\mathcal{L} & \dots & \nabla_{\boldsymbol{W}_{D,D}}\mathcal{L}
    \end{bmatrix} \\
    &= 
    \begin{bmatrix}
    (\nabla_{\boldsymbol{W\psi}_i}\mathcal{L})^T (\nabla_{\boldsymbol{W}_{1,1}}\boldsymbol{W\psi}_i)^T & \dots & (\nabla_{\boldsymbol{W\psi}_i}\mathcal{L})^T (\nabla_{\boldsymbol{W}_{1,D}}\boldsymbol{W\psi}_i)^T \\
    \vdots & & \\
    (\nabla_{\boldsymbol{W\psi}_i}\mathcal{L})^T (\nabla_{\boldsymbol{W}_{D,1}}\boldsymbol{W\psi}_i)^T & \dots & (\nabla_{\boldsymbol{W\psi}_i}\mathcal{L})^T (\nabla_{\boldsymbol{W}_{D,D}}\boldsymbol{W\psi}_i)^T
    \end{bmatrix} \label{eq:update-rule-full-general-gradient}\\
    &= 
    \begin{bmatrix}
    (\nabla_{\boldsymbol{W\psi}_i}\mathcal{L})^T_1 \boldsymbol{\psi}_{i,1} & \dots & (\nabla_{\boldsymbol{W\psi}_i}\mathcal{L})^T_1 \boldsymbol{\psi}_{i,D} \\
    \vdots & & \\
    (\nabla_{\boldsymbol{W\psi}_i}\mathcal{L})^T_D \boldsymbol{\psi}_{i,1} & \dots & (\nabla_{\boldsymbol{W\psi}_i}\mathcal{L})^T_D \boldsymbol{\psi}_{i,D}
    \end{bmatrix} \label{eq:update-rule-full-specific-gradient} \\
    &= (\nabla_{\boldsymbol{W\psi}_i}\mathcal{L}) \boldsymbol{\psi}_i^T
\end{align}
The step between \cref{eq:update-rule-full-general-gradient} and \cref{eq:update-rule-full-specific-gradient} becomes clearer by studying the gradient $\nabla_{\boldsymbol{W}_{j,k}}\boldsymbol{W\psi}_i$:
\begin{align}
    (\nabla_{\boldsymbol{W}_{j,k}}\boldsymbol{W\psi}_i)_l &= \nabla_{\boldsymbol{W}_{j,k}} \sum_{m=1}^D \boldsymbol{W}_{l,m} \boldsymbol{\psi}_{i,m} = \begin{cases} \boldsymbol{\psi}_{i,k}, & \text{if $j$ and $l$ are the same} \\
    0, & \text{otherwise}
    \end{cases}
\end{align}
Thus, the full gradient $\nabla_{\boldsymbol{W}_{j,k}}\boldsymbol{W\psi}_i \in \mathbb{R}^{D}$ is a vector with all zeros except in component $j$ where it is $\boldsymbol{\psi}_{i,k}$. Thus, the dot product $(\nabla_{\boldsymbol{W\psi}_i}\mathcal{L})^T (\nabla_{\boldsymbol{W}_{j,k}}\boldsymbol{W\psi}_i)^T$ will only have a single non-zero term corresponding to $(\nabla_{\boldsymbol{W\psi}_i}\mathcal{L})_j \boldsymbol{\psi}_{i,k}$.
%

Similarly, for $\nabla_{\boldsymbol{\psi}_i}\mathcal{L}$, we have:
\begin{align}
    \nabla_{\boldsymbol{\psi}_i}\mathcal{L} &= 
    \begin{bmatrix}
    \nabla_{\boldsymbol{\psi}_{i,1}}\mathcal{L} \\
    \vdots \\
    \nabla_{\boldsymbol{\psi}_{i,D}}\mathcal{L} 
    \end{bmatrix} \\
    &= 
    \begin{bmatrix}
    (\nabla_{\boldsymbol{W\psi}_i}\mathcal{L})^T (\nabla_{\boldsymbol{\psi}_{i,1}}\boldsymbol{W\psi}_i)^T  \\
    \vdots \\
    (\nabla_{\boldsymbol{W\psi}_i}\mathcal{L})^T (\nabla_{\boldsymbol{\psi}_{i,D}}\boldsymbol{W\psi}_i)^T
    \end{bmatrix} \\
    &= 
    \begin{bmatrix}
    (\nabla_{\boldsymbol{W\psi}_i}\mathcal{L})^T [ \boldsymbol{W}_{1,1}, \dots,  \boldsymbol{W}_{D,1}]^T \\
    \vdots  \\
    (\nabla_{\boldsymbol{W\psi}_i}\mathcal{L})^T  [ \boldsymbol{W}_{1,D}, \dots,  \boldsymbol{W}_{D,D}]^T \\
    \end{bmatrix} \\
     &= \boldsymbol{W}^T (\nabla_{\boldsymbol{W\psi}_i}\mathcal{L})
\end{align}
Now, we can express the update rule:
\begin{align}
    \log \boldsymbol{\alpha}_i^{(t+1)} &= \boldsymbol{W}^{(t+1)}\boldsymbol{\psi}_i^{(t+1)} \\
    &= (\boldsymbol{W} - \eta \nabla_{\boldsymbol{W}}\mathcal{L})(\boldsymbol{\psi}_i - \eta \nabla_{\boldsymbol{\psi}_i}\mathcal{L}) \\
    &= (\boldsymbol{W} - \eta (\nabla_{\boldsymbol{W\psi}_i}\mathcal{L}) \boldsymbol{\psi}_i^T)(\boldsymbol{\psi}_i - \eta \boldsymbol{W}^T (\nabla_{\boldsymbol{W\psi}_i}\mathcal{L})) \\
    &= \boldsymbol{W}\boldsymbol{\psi}_i - \eta\boldsymbol{W}\boldsymbol{W}^T (\nabla_{\boldsymbol{W\psi}_i}\mathcal{L}) \\
    &- \eta (\nabla_{\boldsymbol{W\psi}_i}\mathcal{L}) \boldsymbol{\psi}_i^T\boldsymbol{\psi}_i + \eta^2 (\nabla_{\boldsymbol{W\psi}_i}\mathcal{L}) \boldsymbol{\psi}_i^T\boldsymbol{W}^T (\nabla_{\boldsymbol{W\psi}_i}\mathcal{L}) \\
    &= \boldsymbol{W}\boldsymbol{\psi}_i - \eta(\boldsymbol{W}\boldsymbol{W}^T + \boldsymbol{\psi}_i^T(\boldsymbol{\psi}_i - \eta \boldsymbol{W}^T (\nabla_{\boldsymbol{W\psi}_i}\mathcal{L}))\boldsymbol{I} ) (\nabla_{\boldsymbol{W\psi}_i}\mathcal{L}) \\
    &= \boldsymbol{W}\boldsymbol{\psi}_i - \eta \boldsymbol{T}_i \nabla_{\boldsymbol{W\psi}_i}\mathcal{L} 
\end{align}
which is the same as in Equation~\ref{eq:ip_update_rule}.


\textbf{Scalar Weight.} With a scalar weight $w$, we have $\log \boldsymbol{\alpha}_i^{(t+1)} = w^{(t+1)}\boldsymbol{\psi}_i^{(t+1)}$ and:
\begin{align}
        w^{(t+1)} \leftarrow w - \eta \nabla_{w}\mathcal{L} \\
        \boldsymbol{\psi}_i^{(t+1)} \leftarrow \boldsymbol{\psi}_i - \eta \nabla_{\boldsymbol{\psi}_i}\mathcal{L}
\end{align}
Thus, to derive the update rule for $\log \boldsymbol{\alpha}_i^{(t+1)}$ we need to first derive $\nabla_{w}\mathcal{L}$ and $\nabla_{\boldsymbol{\psi}_i}\mathcal{L}$. We have:
\begin{align}
    \nabla_{w}\mathcal{L} &= (\nabla_{w\boldsymbol{\psi}_i}\mathcal{L})^T (\nabla_{w}w\boldsymbol{\psi}_i) = (\nabla_{w\boldsymbol{\psi}_i}\mathcal{L})^T \boldsymbol{\psi}_i
\end{align}
and:
\begin{align}
    \nabla_{\boldsymbol{\psi}_i}\mathcal{L} &= 
    \begin{bmatrix}
    \nabla_{\boldsymbol{\psi}_{i,1}}\mathcal{L} \\
    \vdots \\
    \nabla_{\boldsymbol{\psi}_{i,D}}\mathcal{L} 
    \end{bmatrix} \\
    &= 
    \begin{bmatrix}
    (\nabla_{w\boldsymbol{\psi}_i}\mathcal{L})^T (\nabla_{\boldsymbol{\psi}_{i,1}}w\boldsymbol{\psi}_i)^T  \\
    \vdots \\
    (\nabla_{w\boldsymbol{\psi}_i}\mathcal{L})^T (\nabla_{\boldsymbol{\psi}_{i,D}}w\boldsymbol{\psi}_i)^T
    \end{bmatrix} \\
    &= 
    \begin{bmatrix}
    (\nabla_{w\boldsymbol{\psi}_i}\mathcal{L})^T [ w, 0, \dots,  0]^T \\
    \vdots  \\
    (\nabla_{w\boldsymbol{\psi}_i}\mathcal{L})^T  [ 0, \dots, 0, w]^T \\
    \end{bmatrix} \\
     &= w (\nabla_{w\boldsymbol{\psi}_i}\mathcal{L})
\end{align}
Now, we can express the update rule:
\begin{align}
    \log \boldsymbol{\alpha}_i^{(t+1)} &= w^{(t+1)}\boldsymbol{\psi}_i^{(t+1)} \\
    &= (w - \eta \nabla_{w}\mathcal{L})(\boldsymbol{\psi}_i - \eta \nabla_{\boldsymbol{\psi}_i}\mathcal{L}) \\
    &= (w - \eta (\nabla_{w\boldsymbol{\psi}_i}\mathcal{L})^T \boldsymbol{\psi}_i)(\boldsymbol{\psi}_i - \eta w (\nabla_{w\boldsymbol{\psi}_i}\mathcal{L})) \\
    &= w\boldsymbol{\psi}_i - \eta w^2 (\nabla_{w\boldsymbol{\psi}_i}\mathcal{L}) \\
    &- \eta ((\nabla_{w\boldsymbol{\psi}_i}\mathcal{L})^T \boldsymbol{\psi}_i)\boldsymbol{\psi}_i + \eta^2 w((\nabla_{w\boldsymbol{\psi}_i}\mathcal{L})^T \boldsymbol{\psi}_i) (\nabla_{w\boldsymbol{\psi}_i}\mathcal{L}) \\
    &= w\boldsymbol{\psi}_i - \eta(w^2 - \eta w (\nabla_{w\boldsymbol{\psi}_i}\mathcal{L})^T \boldsymbol{\psi}_i) (\nabla_{w\boldsymbol{\psi}_i}\mathcal{L}) - \eta ((\nabla_{w\boldsymbol{\psi}_i}\mathcal{L})^T \boldsymbol{\psi}_i)\boldsymbol{\psi}_i \\
    &= w\boldsymbol{\psi}_i - \eta w(w - \eta \nabla_{w}\mathcal{L})(\nabla_{w\boldsymbol{\psi}_i}\mathcal{L}) - \eta (\nabla_{w}\mathcal{L})\boldsymbol{\psi}_i \\
    &= w\boldsymbol{\psi}_i - \eta (ww^{(t+1)}(\nabla_{w\boldsymbol{\psi}_i}\mathcal{L}) + (\nabla_{w}\mathcal{L})\boldsymbol{\psi}_i )
\end{align}
Thus, with a scalar weight $w$ the update rule takes on a slightly different form compared to the full matrix $\boldsymbol{W}$. One interpretation is that the standard gradients $\nabla_{w\boldsymbol{\psi}_i}\mathcal{L}$ are still scaled, but now by $w^{(t)}w^{(t+1)}$ and furthermore, differently from the full matrix case, the gradients are also translated by $(\nabla_{w}\mathcal{L})\boldsymbol{\psi}_i$.

\end{document}